\documentclass[12pt, draftclsnofoot, onecolumn]{IEEEtran}
\usepackage{stmaryrd}
\usepackage{amsmath}
\usepackage{amssymb}
\usepackage{psfrag}
\usepackage{graphicx}
\usepackage{epstopdf}
\usepackage{todonotes}
\usepackage{multirow}
\usepackage{booktabs}
\usepackage{siunitx}
\usepackage{cite}
\usepackage{algorithmic}
\usepackage{algorithm}
\usepackage{svg}
\usepackage{mathtools}
\usepackage{hyperref}

\usepackage{float}
\ifCLASSOPTIONcompsoc
\usepackage[caption=false,font=normalsize,labelfont=sf,textfont=sf]{subfig}
\else
\usepackage[caption=false,font=footnotesize]{subfig}
\fi
\newtheorem{definition}{Definition}
\newtheorem{theorem}{Theorem}
\newtheorem{lemma}{Lemma}
\begin{document}
%
\title{Harnessing Tensor Structures -- Multi-Mode Reservoir Computing and Its Application in Massive MIMO}

\author{Zhou Zhou, Lingjia Liu, and Jiarui Xu
\thanks{Z. Zhou, L. Liu, and J. Xu are with ECE Department at Virginia Tech. The corresponding author is L. Liu (ljliu@ieee.org).}
}
%



\maketitle

\begin{abstract}
In this paper, we introduce a new neural network (NN) structure, multi-mode reservoir computing (Multi-Mode RC).
It inherits the dynamic mechanism of RC and processes the forward path and loss optimization of the NN using tensor as the underlying data format. 
Multi-Mode RC exhibits less complexity compared with conventional RC structures (e.g. single-mode RC) with comparable generalization performance. 
Furthermore, we introduce an alternating least square-based learning algorithm for Multi-Mode RC as well as conduct the associated theoretical analysis. 
The result can be utilized to guide the configuration of NN parameters to sufficiently circumvent over-fitting issues. 
As a key application, we consider the symbol detection task in multiple-input-multiple-output (MIMO) orthogonal-frequency-division-multiplexing (OFDM) systems with massive MIMO employed at the base stations (BSs). 
Thanks to the tensor structure of massive MIMO-OFDM signals, our online learning-based symbol detection method generalizes well in terms of bit error rate even using a limited online training set. Evaluation results suggest that the Multi-Mode RC-based learning framework can efficiently and effectively combat practical constraints of wireless systems (i.e. channel state information (CSI) errors and hardware non-linearity) to enable robust and adaptive learning-based communications over the air.
\end{abstract}


%
\IEEEpeerreviewmaketitle

\section{Introduction}
Massive multiple-input multiple-output (MIMO) is an essential physical layer technique for the 5th generation cellular networks (5G) \cite{marzetta2015massive}. 
By employing a large number of antennas at base stations (BSs), a ``favorable propagation'' channel condition can be achieved. 
It allows inter-user interference being effectively eliminated via fairly simple linear precoding or receiving methods, e.g., conjugate beamforming for downlink, or matched filtering for uplink \cite{ gao2015massive}. 

However, the deployment of massive MIMO in practical systems encounters several implementation constraints. 
Primarily, for the sake of achieving the promised benefits by massive MIMO, highly accurate channel state information (CSI) is needed \cite{truong2013effects}.
On the other hand, CSI with high precision is challenging to be obtained due to the low received signal-to-noise (SNR) before beamforming/precoding, as well as the limited pilot symbols defined in modern cellular networks due to control overhead~\cite{r2020}. 
Furthermore, theoretical analysis of massive MIMO systems usually assume ideal linearity and pleasing noise figures requiring exceptionally high costs on radio frequency (RF) and mixed analog-digital components~\cite{vieira2014flexible}. This ends up with a compromise on the hardware selection which introduces imperfectness (e.g. dynamic non-linearity) to the transmission link. The resulting non-linearity, on the other hand, leads to waveform distortion and thereby diminishes the transmission reliability, which is challenging to be analytically tackled using model-based approaches.

Symbol detection is a critical stage in wireless communications. It is a process that accomplishes miscellaneous interference cancellation at receivers, such as inter-symbol, inter-stream, and inter-user inference, etc. Due to the potential model mismatch from the non-linearity caused by low-cost hardware devices, standard model-based signal processing approaches are no longer effective. With the advent of deep neural networks, there are growing interests in using neural networks (NNs) to handle the model mismatch~\cite{r2020}. In general, NN-based framework aims to compensate for the model mismatch through the non-linearity of NNs. This recent awareness of bridging learning-based approaches to the symbol detection task in massive MIMO systems has posed the following conceptual discussions in the NN design.
\begin{itemize}
\item Curse of Antenna Dimensionality:
Since the input, hidden, and output layers of NNs are often configured as the same scale as the antennas to jointly extract and process spatial and time-frequency features, the growth of antenna numbers inevitably lead to the increase of the volume of underlying NN coefficients. As NNs essentially learn the underlying statistics of data, the corresponding increase in the parameter dimensionality often imposes an exponential need on the training data set to offer a reasonable generalization result. However, the availability of the training data for cellular networks (e.g.,4G or 5G) especially the online ones is extremely limited due to the associated control overhead~\cite{zhou2020learning}. Furthermore, the computational complexity is also evinced with an exponential relation to the NN scale. Learning neural weights through generic back-propagation can result in large computational complexities leading to severe processing delays which are not desirable especially for delay-sensitive applications. 

\item Blessings of Antenna Dimensionality: On the other hand, a large number of antennas is able to offer favorable propagation conditions~\cite{marzetta2015massive}. 
Properly leveraging the asymptotic orthogonality of the wireless channels can often result in surprising outcomes, which conversely transforms the curse of dimensionality into ``the blessings of dimensionality''~\cite{bjornson2016massive}. 
These findings align with the measurement concentration phenomena which are widely applied to simplifying machine learning frameworks~\cite{gorban2018blessing}. 
Therefore, to explore learning-based strategies for massive MIMO, we can incorporate inherent structures from the spatial channel as well as time-frequency features from the modulation waveform to the design of NNs, which is ``blessed'' to offer good generalization performance yet under very limited online training.
\end{itemize}

\subsection{Previous Work}
A commonly utilized approach for building learning-based symbol detectors is through unfolding existing optimization-based symbol detection methods to deep NNs, such as DetNet\cite{samuel2019learning} and MMNet\cite{khani2019adaptive}. 
Since this framework is based on using explicit CSI, it usually suffers from performance drop or requires extensive hyper-parameters tuning when CSI is not perfect. Meanwhile, the resulting ``very deep neural networks'' are extremely demanding in computational resources which hinders their applications in practical scenarios. Alternatively, implicit CSI can be utilized to circumvent the above mentioned training issues. For example, \cite{Ye2018MIMO} introduced a deep feedforward NNs for symbol detection in single-input-single-output (SISO) OFDM systems. Due to its independence from channel models, this approach can equalize the channel with nonlinear distortion (power amplifier (PA)). 
However, this method uses a less-structured deep NN which is yet too complicated to train in practice, since the guaranteed generalization performance is based on extensive training over large datasets that are impossible to obtain in over-the-air scenarios. Furthermore, “uncertainty in generalization”~\cite{r2020} will arise if the dataset used for training the underlying NN is not general enough to capture the distribution of data encountered in testing. This is especially true for 5G and Beyond 5G networks that needs to offer reliable service under vastly different scenarios and environments. 

In 4G/5G MIMO-OFDM systems, there exists different operation modes with link adaptation, rank adaptation, and scheduling on a subframe basis~\cite{LiuMIMO}. Therefore, it is challenging, to adopt a complete offline training-based approach. Rather, it is critical to design an online NN-based approach to conduct symbol detection in each subframe only using the limited training symbols that are present in that particular subframe. In this way, the online-learning-based approach can be adaptive and robust to the change of operation modes, channel distributions, and environments. On the other hand, conducting effective and efficient learning only through the limited training symbols within a subframe is extremely challenging. To achieve this goal, more structural knowledge of the wireless channel and modulation waveform need to be incorporated as inductive priors to the NNs to significantly relieve the training overhead in each subframe basis\cite{zhou2020learning2,zhou2020learning}. Reservoir computing (RC) and its deep version have been introduced for the MIMO-OFDM symbol detection task in \cite{mosleh2017brain,zhou2019,zhou2020deep,zhou2020rcnet} to achieve learning on a subframe basis. 
To be specific, \cite{mosleh2017brain} is the first work using a vanilla RC structure to conduct MIMO-OFDM symbol detection, where the input and output are defined in the time domain. It shows this simple approach can achieve good symbol detection performance in short memory channels with limited training. \cite{zhou2019,zhou2020deep,zhou2020rcnet} extended the RC-based symbol detection framework by adding units in width and deepth to handle channel with long taps as well as more severe non-linearity. Experiments show that the extended RC framework -- RCNet can effectively compensate for the distortion caused by non-linear components in wireless systems as well as mitigate miscellaneous interference merely from receiver side using training dataset only from each subframe. 

\subsection{Contributions}
In this paper, we consider the uplink symbol detection in a massive MIMO system with OFDM waveform using a ``subframe by subframe'' learning framework. Uplink transmission is a typical low SNR scenario since mobile terminals often use relatively low transmission powers, and the RC framework has not yet been investigated under the scope of the massive MIMO systems. Being able to conduct receive processing -- symbol detection on a subframe basis is extremely important for robust and adaptive communications in the 5G and beyond 5G massive MIMO networks.
By referring to the multi-dimensional feature of massive MIMO signals (e.g. elevation and azimuth directions in the spatial domain, the time and frequency domain), we are motivated to incorporate this tensor structure into our symbol detection NN. Although the concept of tensor-driven NNs has been studied before, such as Tensorized NNs in \cite{novikov2015tensorizing}, where NN weights are formulated as a tensor-train decomposition \cite{oseledets2011tensor}, and CANDECOMP/PARAFAC (CP) decomposition characterized convolutional layers for learning-acceleration from \cite{lebedev2015speeding}, our strategy is different to these techniques as tensor is utilized as the forward path data structure rather than NN coefficients. In its application to massive MIMO systems, instead of treating the input signal as a vector sequence, we define the received signal as a tensor sequence that is consistent with the intrinsic multiple mode property of the underlying massive MIMO signals. Such a signal processing perspective has been studied in \cite{Cheng15TSP, r16, r17, zhou2017low, zhou2016channel, zhou2019fd} to solve conventional massive MIMO channel estimation problems. However, a more accurate explicit CSI does not sufficiently lead to an improvement of the transmission reliability, since symbol detection is conducted on a separate stage without knowledge of the channel estimation errors. Our introduced method is to directly demodulate symbols avoiding the intermediate channel estimation stage. 
More importantly, our method accomplishes the symbol detection by using training dataset only from each subframe.
The resulting multi-mode processing framework can in general be extended to process any other tasks with tensor structured sequence, such as video, social networks, and recommendation systems, etc.

We name our introduced RC-based NN structure as ``multi-mode reservoir computing'' (Multi-mode RC), as it inherits the dynamic mechanism of RC and processes input-output relation using a tensor format (multi-dimensional array). In our framework, a core-tensor is built as hidden features of the input tensor sequence. Desired output is then obtained through a multi-mode mapping. In terms of tensor algebra, the RC readout is learned through a Tucker decomposition with a deterministic core-tensor thanks to the aforementioned feature extraction. A theoretical analysis is then provided to show the uniqueness of the learned NN coefficients. Our experiments reveal that this uniqueness condition is related to the avoidance of over-fitting issues since it prevents a zero loss value which often results in poor generalization performance on the testing dataset. Compared to single-mode RC, Multi-Mode RC can achieve better symbol detection performance in terms of uncoded bit error rate in the low SNR regime with reduced computational complexity. 
In addition, the introduced method is shown to be effective to combat extreme waveform distortion, e.g. applying one-bit analog to digital converter (ADC) as the receiving quantization.
The remainder of this paper is organized as follows: In Sec. \ref{prelim}, we briefly introduce math foundations of tensor and the background of reservoir computing which are utilized to build the concept of Multi-Mode RC in Sec. \ref{muti_mode_rc}. In Sec. \ref{application_mimo_ofdm}, the application of Multi-Mode RC to massive MIMO-OFDM symbol detection is discussed. Sec. \ref{Evaluations} evaluates the performance of Multi-Mode RC as opposed to existing symbol detection strategies in massive MIMO-OFDM systems. The conclusion and future research directions are outlined in Sec. \ref{Conclusion}.

\section{Preliminary}
\label{prelim}
This section provides an overview of basic tensor algebra and reservoir computing which will be used in the rest of this paper as the methodology development. In our math notations: scalars, vector, matrix, and tensor are denoted by lowercase letters, boldface lowercase letters, boldface uppercase letters, and boldface Euler script letters respectively, e.g., $x$, $\boldsymbol x$, $\boldsymbol X$ and $\boldsymbol {\mathcal X}$. 

\subsection{Tensor Algebra}
Tensor is an algebraic generalization to matrix. A tensor represents a multidimensional array, where the mode of a tensor is the number of dimensions, also known as ways and orders\cite{kolda2009tensor}. The $(i_1, i_2, \cdots, i_N)$th element of a $N$-mode tensor, or namely a $N$th order tensor, ${\boldsymbol {\mathcal X}} \in \mathbb{C}^{I_{1} \times I_{2} \times \cdots \times I_{N}}$, is denoted as $x_{i_1,i_2,\cdots, i_N}$, where indices range from $1$ to their capital versions. 

\begin{figure}[h]
    \centering
    \includegraphics[width = 1\linewidth]{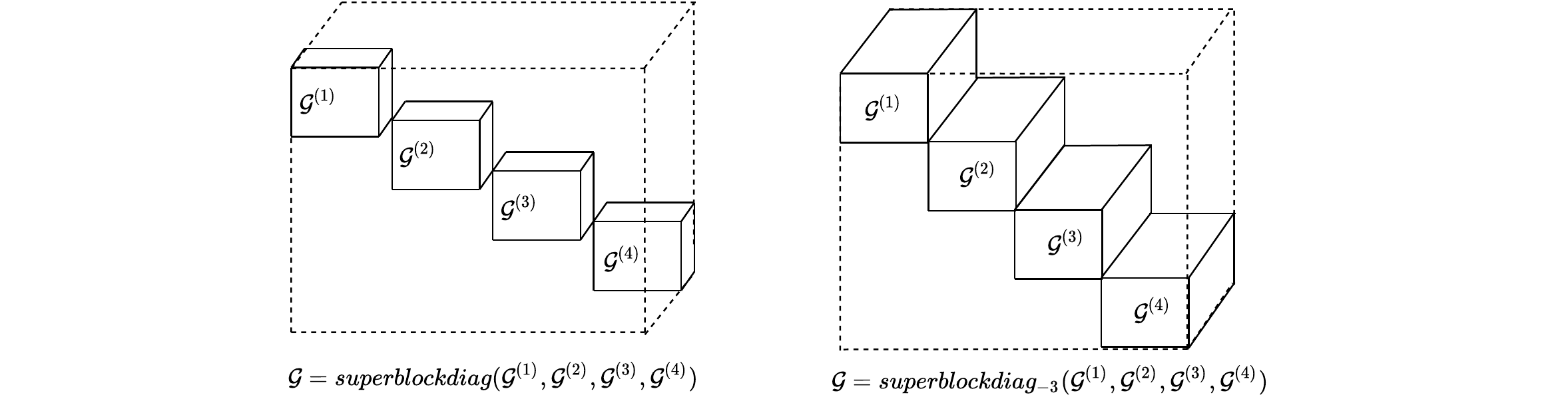}
    \caption{Illustration of the tensor diagonalization of 3-mode tensors: The left hand-side represents a super diagonal tensor ${\text{superblockdiag}}({\boldsymbol {\mathcal G}}^{(1)}, {\boldsymbol {\mathcal G}}^{(2)}, {\boldsymbol {\mathcal G}}^{(3)}, {\boldsymbol {\mathcal G}}^{(4)})$, the right hand side represents a super diagonal tensor only on the first two modes, i.e. ${\text{superblockdiag}}_{-3}({\boldsymbol {\mathcal G}}^{(1)}, {\boldsymbol {\mathcal G}}^{(2)}, {\boldsymbol {\mathcal G}}^{(3)}, {\boldsymbol {\mathcal G}}^{(4)})$, where the diagonal elements are ${\boldsymbol {\mathcal G}}^{(1)}$, ${\boldsymbol {\mathcal G}}^{(2)}$, ${\boldsymbol {\mathcal G}}^{(3)}$ and ${\boldsymbol {\mathcal G}}^{(4)}$.}
    \label{super_tensor}
\end{figure}

By following matrix conventions, $rank({\boldsymbol X})$ represents the rank of the matrix $\boldsymbol X$. ${\boldsymbol X}^T$, ${\boldsymbol X}^H$ and ${\boldsymbol X}^+$ respectively stands for the transpose, hermitian transpose, and Moore–Penrose pseudoinverse of the matrix $\boldsymbol X$. Analogously, the tensor transpose of a tensor $\boldsymbol {\mathcal X}$ is denoted as ${\boldsymbol {\mathcal X}}^{T_{\Pi}}$ which means the $i$th mode of ${\boldsymbol {\mathcal X}}^{T_{\Pi}}$ correspond to the mode numbered as $\Pi(i)$ of $\boldsymbol {\mathcal X}$, where $\Pi$ is a permutation on set $\{1,2,..,N\}$. Moreover, ${\text{blockdiag}}({{\boldsymbol A}_1}, \cdots, {\boldsymbol A}_N)$ represents stacking ${{\boldsymbol A}_1}, \cdots, {\boldsymbol A}_N$ as a block-diagonal matrix.  We denote ${\text{superblockdiag}}(·)$ as a super-diagonal tensor by stacking its tensor arguments as illustrated in Fig. \ref{super_tensor}. $\text{superblockdiag}_{-n}(\cdot)$ forms a super diagonal tensor except on mode $n$.

The definition of the mode-$n$ matricization of a tensor $\boldsymbol {\mathcal X}$ is denoted as $\mathbf{X}_{(n)}$,
where $\left(i_{1}, i_{2}, \ldots, i_{N}\right)$ of ${\boldsymbol {\mathcal X}} \in \mathbb{C}^{I_{1} \times I_{2} \times \cdots \times I_{N}}$ maps to the $\left(i_{n}, j\right)$ entry of matrix $\mathbf{X}_{(n)}\in {\mathbb C}^{I_n\times I_{-n}}$, where $I_{-n}:=\prod_{k \neq -n}I_{k} $. According to\cite{tensorly},
\begin{equation}
\label{unfold_order}
j:=1+\sum_{k=1 \atop k \neq n}^{N}\left(i_{k}-1\right) J_{k} \quad \text { with } \quad J_{k}=\prod_{m=k+1 \atop m \neq n}^{N} I_{m}.
\end{equation}
The $n$-mode product of a tensor $\boldsymbol {\mathcal X}$ with a matrix ${\boldsymbol U}\in {\mathbb C}^{J\times I_n}$ is defined as,
\begin{equation*}
\left({\boldsymbol {\mathcal X}} \times_{n} \mathbf{U}\right)_{i_{1} \ldots i_{n-1} j i_{n+1} \ldots i_{N}}=\sum_{i_{n}=1}^{I_{n}} x_{i_{1} i_{2} \ldots i_{N}} u_{j i_{n}}.
\end{equation*}
The Tucker decomposition of a tensor $\boldsymbol {\mathcal X}$ is defined as
\begin{equation}
\boldsymbol{\mathcal X}=\boldsymbol{\mathcal G} \times_{1} \mathbf{A}^{(1)} \times_{2} \mathbf{A}^{(2)} \cdots \times_{N} \mathbf{A}^{(N)}
\label{tk0}
\end{equation}
where ${\boldsymbol A}^{(n)}$ represents the $n$th factor matrix and $\boldsymbol {\mathcal G}$ is named as the core tensor. Accordingly, the mode-$n$ unfolding of the tensor $\boldsymbol {\mathcal X}$ is given by
\begin{align}
    \label{unfolding_form}
    {\boldsymbol X}_{(n)} = {\boldsymbol A}^{(n)}{\boldsymbol G}_{(n)}({\boldsymbol A}^{(1)}\otimes\cdots{\boldsymbol A}^{n-1}\otimes {\boldsymbol A}^{n+1}\cdots\otimes{\boldsymbol A}^{(N)})^T,
\end{align}
where ${\boldsymbol G}_{(n)}$ is the mode-n unfolding of $\boldsymbol {\mathcal G}$. Note the above unfolding tensor has a reverse order in the Kronecker products of factor matrices which differs to \cite{kolda2009tensor}. This is because we alter the way to pile up the indices of unfolding tensors according to (\ref{unfold_order}).

\begin{figure}
    \centering
    \includegraphics[width = 1\linewidth]{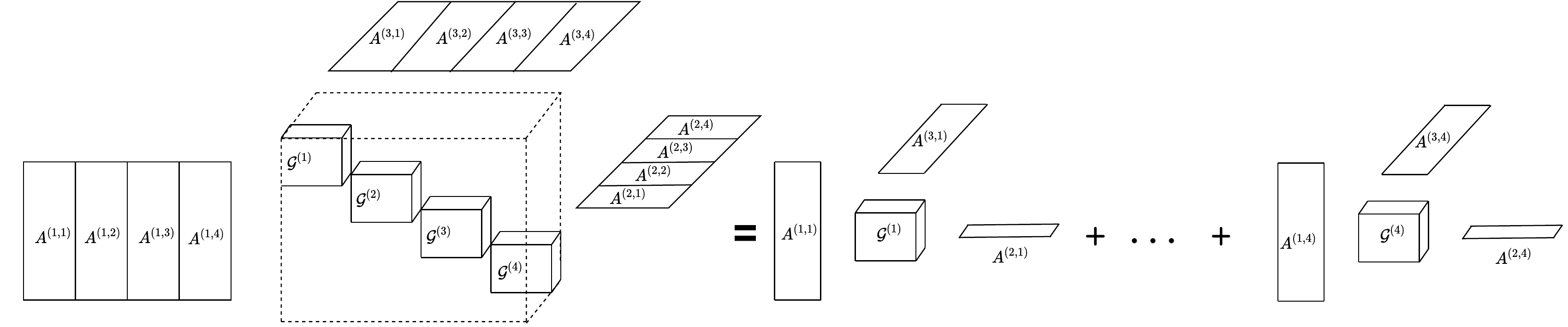}
    \caption{Illustration for a Tucker decomposition of a three mode tensor: Core tensor and factor matrices are with four partitions.}
    \label{tucker_decomposition}
\end{figure}

We now consider a super diagonal core tensor $\boldsymbol{\mathcal G}$ with $K$ blocks, i.e., 
\begin{align*}
    { {\boldsymbol {\mathcal G}}} = \text{superblockdiag}({\boldsymbol{\mathcal G}}^{(1)},{\boldsymbol{\mathcal G}}^{(2)},\cdots, {\boldsymbol{\mathcal G}}^{(K)} )
\end{align*}
where ${\boldsymbol{\mathcal G}}^{(k)}\in {\mathbb C}^{I_1^{(k)}\times I_2^{(k)}\times\cdots I_N^{(k)}}$ and a matrix ${\boldsymbol A}^{(n)}$ being partitioned as 
\begin{align*}
[{\boldsymbol A}^{(n,1)}, {\boldsymbol A}^{(n,2)},\cdots, {\boldsymbol A}^{(n,K)}].
\end{align*}
Accordingly, the resulting n-mode product between $\boldsymbol{\mathcal G}$ and ${\boldsymbol A}^{(n)}$ can be written in terms of a super diagonal tensor except on the $n$th mode:
\begin{align*}
{{\boldsymbol {\mathcal G}}}\times_n {\boldsymbol A}^{(n)}
= \text{superblockdiag}_{-n}({\boldsymbol{\mathcal G}}^{(1)}\times_n{\boldsymbol A}^{(n,1)},{\boldsymbol{\mathcal G}}^{(2)}\times_n{\boldsymbol A}^{(n,2)},\cdots, {\boldsymbol{\mathcal G}}^{(K)}\times_n{\boldsymbol A}^{(n,K)}).
\end{align*}
When we assume the core tensor of $\boldsymbol {\mathcal X}$ is super-diagonal, the Tucker decomposition defined in (\ref{tk0}) can be alternatively expressed in terms of a summation of sub-Tucker decompositions:
\begin{align}
\label{partition_tucker}
\boldsymbol{\mathcal X}=\sum_{k=1}^K\boldsymbol{\mathcal G}^{(k)} \times_{1} \mathbf{A}^{(1,k)} \times_{2} \mathbf{A}^{(2,k)} \cdots \times_{N} \mathbf{A}^{(N,k)}.
\end{align} An illustration for Tucker decomposition of a three mode tensor is depicted in Fig. \ref{tucker_decomposition}.

\subsection{Reservoir Computing}

Reservoir computing (RC) is defined as a framework for computation by using memory units. The `reservoir' is composed of nonlinear components and recurrent loops, where the non-linearity allows RC to process complex problems and the recurrent loops enable RC with memory. The `computing' is achieved by reading out the states in the reservoir through learned NN layers. The training of this framework is conducted only on the readout layers which fundamentally circumvents gradient vanishing/explosion issues in back-propagation through time thanks to the fixed reservoir dynamics.

A vanilla discrete-time realization of RC is characterized by a state equation and an output equation. The state equation is formulated with time index $t$ by,
\begin{align}
\label{rec_layer}
{\boldsymbol s}(t+1) = \sigma\left({\boldsymbol W}_{tran}
\begin{bmatrix}
{\boldsymbol s}(t)\\
{\boldsymbol y}(t)
\end{bmatrix}\right)
\end{align}
where $\sigma$ is a nonlinear function, ${\boldsymbol s}(t)$ is a vector representing the internal reservoir state, ${\boldsymbol y}(t)$ is the input vector, and ${\boldsymbol W}_{tran}$ stands for the reservoir weight matrix which is often chosen with a spectral radius smaller than 1 in order to asymptotically reduce any impacts from initial states. The output equation is simply treated as
\begin{align}
\label{out_layer}
{\boldsymbol z}(t) = {\boldsymbol W}_{out}
\begin{bmatrix}
{\boldsymbol s}(t)\\
{\boldsymbol y}(t)
\end{bmatrix},
\end{align}
where ${\boldsymbol W}_{out}$ is the output weight matrix and ${\boldsymbol z}(t)$ stands for the output. As we can see, the output is with a skip-connection to the input which assimilates to a residual arrangement \cite{NEURIPS2019_5857d68c}. 

\section{Multi-Mode Reservoir Computing}
\label{muti_mode_rc}
In this section, we introduce the framework of Multi-Mode RC. It processes a sequence-in and sequence-out task, where the time sequences are configured with more than one explicit modes, i.e., the input sequence is formulated as ${\boldsymbol Y}(t)$ or ${\boldsymbol {\mathcal Y}}(t)$ rather than a scalar-wise sequence $y(t)$ or a vector-wise one ${\boldsymbol y}(t)$.
\subsection{Two-Mode Reservoir Computing}
For ease of discussion, we begin by considering a two-mode RC. The architecture is comprised of a recurrent module, a feature queue, and an output mapping.

{\textbf{Recurrent Module}}: A recurrent module maps input sequence ${\boldsymbol Y}(t)\in {\mathbb C}^{N_{in-1}\times N_{in-2}}$ to a state sequence ${\boldsymbol S}(t) \in {\mathbb C}^{N_s \times N_s}$, where $N_s$ represents the number of neurons defined on each mode of ${\boldsymbol S}(t)$\footnote{For simplicity, we assume ${\boldsymbol S}(t)$ as a square matrix. In general, it also can be designed as a non-square matrix. Meanwhile, the size of $N_s$ is configured through experiments in order to maintain a balance between overfitting and underfitting according to the datasets. The RC structure based on non-square ${\boldsymbol S}(t)$ is left as our future work.}. Therefore, the total number of neurons is $N_s^2$. To equally obtain observations from the row-space and column-space of ${\boldsymbol Y}(t)$, we define the recurrent equation as,  
\begin{equation}
\begin{aligned}
{\boldsymbol {S}}(t+1) &= \sigma\left({\boldsymbol W}_{tran-1}
\begin{bmatrix}
{\boldsymbol {S}}(t),& {\boldsymbol O}\\
{\boldsymbol O},& {\boldsymbol {\tilde Y}}(t)
\end{bmatrix}{\boldsymbol W}_{tran-2}^T\right)
\end{aligned}
\label{state_equation_2_mode}
\end{equation}
where 
\begin{align*}
{\boldsymbol {\tilde Y}}(t) = {\text{blockdiag}}({\boldsymbol Y}(t), {\boldsymbol Y}(t-1), \cdots, {\boldsymbol Y}(t-T')),
\end{align*}
$T'$ is a hyper-parameter representing the length of input window, $\sigma$ is a non-linear function, ${\boldsymbol W}_{tran-1} \in {\mathbb C}^{N_s \times (N_s+T'N_{in-1})}$, ${\boldsymbol W}_{tran-2} \in {\mathbb C}^{N_s \times (N_s+T'N_{in-2})}$ are reservoir weight matrices applied on the row and column spaces respectively. Note that (\ref{state_equation_2_mode}) also can be written as a sum as the form of (\ref{partition_tucker}). Accordingly, the state equation (\ref{state_equation_2_mode}) can be regarded as an extension of the standard state equation (\ref{rec_layer}) by incorporating independent mappings into the row and column spaces of state ${\boldsymbol S}(t)$ and input ${\boldsymbol Y}(t)$. It also can be equivalently written via the form of (\ref{rec_layer}) through vectorizing the matrix-wise state and input. Rather than directly applying vectorized state and input to the standard RC, our introduced approach preserves the multi-mode feature of the input signal, where the advantages of using this strategy will be discussed in the analysis and evaluation sections of this paper. 

{\textbf{Feature Queue}:}
Our definition of a feature queue ${\boldsymbol G}(t)$ is a queue of sequence, i.e., at a given time $t$, the sample ${\boldsymbol G}(t)$ is a queue which is stacked up by current state sample ${\boldsymbol S}(t)$ and input sample ${\boldsymbol {\tilde Y}}(t)$. We opt for a simple formulation and use diagonalizing operation to write the queue as follows, 
\begin{align}
    {\boldsymbol G}(t) = {\text{blockdiag}}({{\boldsymbol S}}(t), {{\boldsymbol S}}^T(t), {\tilde{\boldsymbol Y}}(t), {\tilde{\boldsymbol Y}}^T(t)).
\end{align}
${{\boldsymbol G}}(t)$ is also called extended state sequence as it is obtained with a skip connection to the input.  The presence of ${\boldsymbol {\tilde Y}}^T(t)$ and ${\boldsymbol S}^T(t)$ is to create a fair treatment on the row and column space of ${\boldsymbol {\tilde Y}}(t)$ and ${\boldsymbol { S}}(t)$.

{\textbf{Output Mapping}}: An output layer ensures the feature queue can be identically mapped back to our desired output size. It is defined as:
\begin{equation}
\label{output_mode2}
\begin{aligned}
{\boldsymbol Z}(t) &= 
{\boldsymbol W}_{out-1}{\boldsymbol G}(t){\boldsymbol W}_{out-2}^T \\
&= {\boldsymbol G}(t)\times_1 {\boldsymbol W}_{out-1}\times_2 {\boldsymbol W}_{out-2}
\end{aligned}
\end{equation}
where ${\boldsymbol W}_{out-1} \in {\mathbb C}^{N_{out-1} \times N_{f-1}}$ and ${\boldsymbol W}_{out-2} \in {\mathbb C}^{N_{out-2} \times N_{f-2}}$; $N_{f-1} := 2N_s+T'(N_{in-1}+N_{in-2})$ and $N_{f-2} := 2N_s+T'(N_{in-1}+N_{in-2})$  respectively represent the size of row and column of ${\boldsymbol G}(t)$; Meanwhile, $N_{out-1}$ and $N_{out-2}$ respectively stand for the size of row and column of ${\boldsymbol Z}(t)$.
    
{\textbf{Loss Function:}} In this paper, the loss function is defined to handle sequence-to-sequence tasks. Given a set of $\{{\boldsymbol Y}_q(t), {\boldsymbol Z}_q(t)\}$ as the input-output pairs for training, where $q$ stands for the batch index,  our objective aims to generate ${\boldsymbol Z}_q(t)$  by using ${\boldsymbol Y}_q(t)$. Therefore, we use
\begin{align}
\label{obj_2modes}
\min_{{\boldsymbol W}_{out-1}, {\boldsymbol W}_{out-2}} \sum_{q=1}^{N_K}\sum_{t=1}^{N_T}\|{\boldsymbol {Z}}_q(t) - {\boldsymbol {G}}_q(t)\times_1{\boldsymbol W}_{out-1}\times_2 {\boldsymbol W}_{out-2}\|_F^2,
\end{align}
where $\|\cdot\|_F$  is the Frobenius norm of a matrix. Although the loss function is simply formulated via a least square framework, it offers a connection between the RC readout learning and alternating least squares (ALS) algorithm which has been widely used and can be easily analyzed in the context of tensor decomposition\cite{de2008decompositions2}. We can further stack ${\boldsymbol Z}_q(t)$ and ${\boldsymbol G}_q(t)$ to $4$-mode tensors along the time axis and the batches to have ${\boldsymbol {\mathcal Z}}\in {\mathbb C}^{N_{out-1} \times N_{out-2}\times N_T\times N_K}$ and ${\boldsymbol {\mathcal G}}\in {\mathbb C}^{N_{f-1} \times N_{f-2}\times N_T\times N_K}$ respectively. Accordingly, the loss objective (\ref{obj_2modes}) can be rewritten in concise way,
\begin{align}
\label{loss_function}
\min_{{\boldsymbol W}_{out-1}, {\boldsymbol W}_{out-2}} \|{\boldsymbol {\mathcal Z}} - {\boldsymbol {\mathcal G}}\times_1{\boldsymbol W}_{out-1}\times_2 {\boldsymbol W}_{out-2}\|_F^2.
\end{align}

In the training stage, we feed a batch of sequences to RC and solve the problem (\ref{loss_function}) using alternating least squares, where ${\boldsymbol W}_{out-1}$ and ${\boldsymbol W}_{out-2}$ are iteratively updated by solving the following matrix-wise least square problems,
\begin{align*}
    {\boldsymbol W}_{out-1} = \arg\min_{{\boldsymbol W}_{out-1}}\|{\boldsymbol Z}_{(1)} - {\boldsymbol W}_{out-1}{\boldsymbol G}_{(1)}( {\boldsymbol W}_{out-2}\otimes{\boldsymbol I}_{N_T}\otimes{\boldsymbol I}_{N_K})^T\|_F\\
    {\boldsymbol W}_{out-2} =  \arg\min_{{\boldsymbol W}_{out2}}\|{\boldsymbol Z}_{(2)} - {\boldsymbol W}_{out-2}{\boldsymbol G}_{(2)}( {\boldsymbol W}_{out-1}\otimes{\boldsymbol I}_{N_T}\otimes{\boldsymbol I}_{N_K})^T\|_F.
\end{align*}
The iterative process continues until a certain stopping criterion is reached. In this ALS formulation, ${\boldsymbol G}_{(1)}$ and ${\boldsymbol G}_{(2)}$ represent the mode-1 and mode-2 unfoldings of tensor $\boldsymbol {\mathcal G}$. However, directly using this ALS calculation often requires large memory resources due to the Kronecker products. Therefore, we introduce an alternative approach to calculate the ALS which is discussed in Appendix. 

Since reaction delays exist in RC systems\cite{zhou2020deep}, we often need to add another parameter $\tau$, namely ``Delay of RC states'' in the loss objective to optimize. Therefore, we have the following augmented loss objective,
\begin{align}
\label{loss_2_mode}
\min_{\tau < \tau_{max}}\min_{{\boldsymbol W}_{out-1}, {\boldsymbol W}_{out-2}} \sum_{q=1}^{N_K}\sum_{t=1}^{N_T}\|{\boldsymbol {Z}}_q(t) - {\boldsymbol {G}}_q(t+\tau)\times_1{\boldsymbol W}_{out-1}\times_2 {\boldsymbol W}_{out-2}\|_F^2,
\end{align}
where $\tau_{max}$ represents the upper bound of $\tau$ to search. Accordingly, the samples of ${\boldsymbol G}_q(t)$ ranging from time index $t=1$ to $t=N_T + \tau_{max}$ are obtained by using $\{{\boldsymbol Y}(t)\}_{t=1}^{N_T}$ with a $\tau_{max}$-length zero paddings at the end as the RC input. At testing stage, the learned $\tau$ are applied to truncate the RC state sequence such that the output sequence becomes $\{{\boldsymbol G}(t+\tau)\times_1 {\boldsymbol W}_{out-1} \times_2 {\boldsymbol W}_{out-2}\}_{t = 1}^{N_T}$, since the output is anticipated as a $N_T$-length sequence. In general, the output sequence can be truncated off more or less samples in order to match the desired sequence of the tasks.  
\begin{table}[]
\centering
\caption{Notations of Multi-Mode RC}
\begin{tabular}{|l|l|l|}
\hline
Notations & Definitions \\ \hline
$T'$ & Input window length \\ \hline
$N$ & Number of signal mode in Multi-Mode RC \\ \hline
$N_{in-n}$ & Number of RC input at mode $n$ \\\hline
$N_{out-n}$ & Number of RC output at mode $n$  \\\hline
$N_{f-n}$ & Feature queue size at mode $n$\\ \hline
$N_T$ & Input and output sequence length  \\\hline
$N_K$ & Training batch size \\\hline
$N_s$ & Number of neurons on each mode of a state tensor\\ \hline
$\tau$ & Delay configuration in RC state response \\ \hline
${\boldsymbol {\mathcal Y}}(t) \in {\mathbb C}^{N_{in1}\times N_{in-2}\times \cdots \times N_{in-N}}$ & A tensor sequence as the input of Multi-Mode RC\\\hline
${\boldsymbol {\mathcal G}}(t) \in {\mathbb C}^{N_{f-1}\times N_{f-2}\times \cdots \times N_{f-N}}$ & A tensor sequence as the internal feature of Multi-Mode RC \\\hline
${\boldsymbol {\mathcal Z}}(t) \in {\mathbb C}^{N_{out-1}\times N_{out-2} \cdots \times N_{out-N}}$ & A tensor sequence as the output of RC \\ \hline
${{\boldsymbol {\mathcal Y}}} \in {\mathbb C}^{N_{in1}\times N_{in2}\times \cdots \times N_{in-N} \times N_T \times N_K}$    &  A higher order tensor by stacking ${\boldsymbol {\mathcal Y}}(t)$ through time samples and batches \\ \hline
${{\boldsymbol {\mathcal G}}} \in {\mathbb C}^{N_{f-1}\times N_{f-2}\times \cdots \times N_{f-N} \times N_T \times N_K}$    &  A higher order tensor by stacking ${\boldsymbol {\mathcal G}}(t)$ through time samples and batches    \\ \hline
${{\boldsymbol {\mathcal Z}}} \in {\mathbb C}^{N_{out-1}\times N_{out-2} \cdots N_{out-N} \times N_T \times N_K}$    &  A higher order tensor by stacking ${\boldsymbol {\mathcal Z}}(t)$ through time samples and batches \\ \hline
\end{tabular}
\end{table}
\subsection{Multi-Mode Reservoir Computing}

\begin{figure}[h]
\centering
\includegraphics[width=1\linewidth]{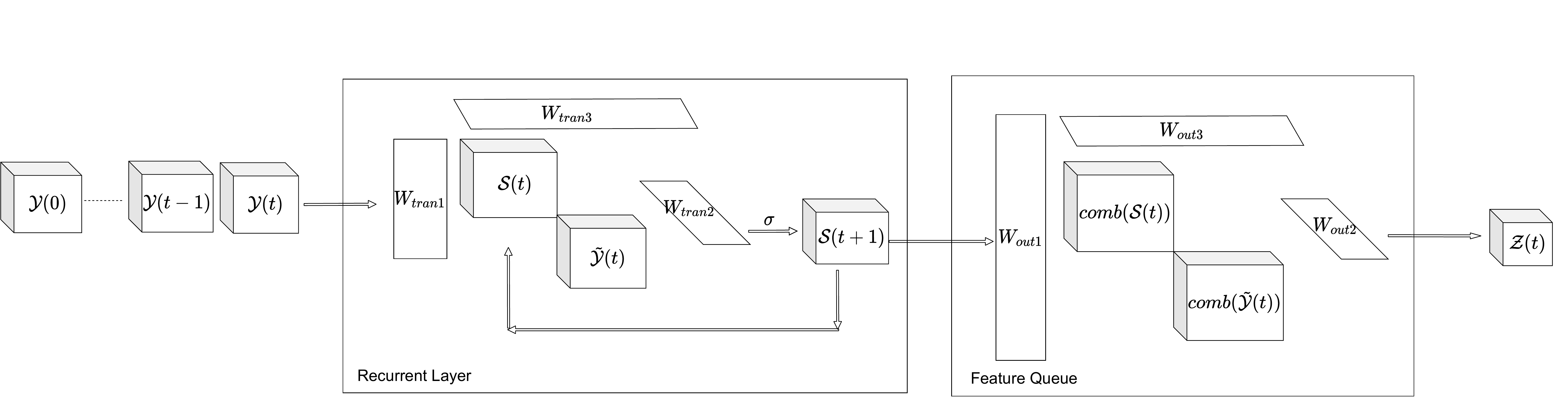}
\caption{Illustration of Three-Mode Reservoir Computing Architecture}
\label{Multi_mode_RC}
\end{figure}

The structure of multi-mode (beyond $2$-mode) reservoir computing is illustrated in Fig. \ref{Multi_mode_RC}. As a general framework of the $2$-mode RC, each component is respectively extended as 
\begin{itemize}
    \item {\textbf{Recurrent Module:}}
    \begin{equation}
    \begin{aligned}
    {\boldsymbol {\mathcal S}}(t+1) &= \sigma({\text{superblockdiag}}({\boldsymbol {\mathcal S}}(t),{\tilde{\boldsymbol {\mathcal Y}}}(t))\times_1 {\boldsymbol W}_{tran-1} \times_2 \cdots \times_N {\boldsymbol W}_{tran-N}) \\
            {\boldsymbol {\mathcal {\tilde Y}}}(t) &= {\text{superblockdiag}}({\boldsymbol {\mathcal Y}}(t), \cdots, {\boldsymbol {\mathcal Y}}(t-T'))
    \end{aligned}
    \end{equation}
    \item {\textbf{Feature Queue}:}
    \begin{equation}
    \begin{aligned}
    &{\boldsymbol {\mathcal G}}(t) = {\text{superblockdiag}}({\text{comb}}({{\boldsymbol {\mathcal S}}}(t)), {\text{comb}}({\tilde{\boldsymbol {\mathcal Y}}}(t))) 
    \end{aligned}
    \label{feature_queue}
    \end{equation}
    where ${\text{comb}}({{\boldsymbol {\mathcal S}}}(t)) := {\text{superblockdiag}}({{\boldsymbol {\mathcal S}}}(t),{{\boldsymbol {\mathcal S}}}^{T_{\Pi_1}}(t), {{\boldsymbol {\mathcal S}}}^{T_{\Pi_2}}(t)\cdots )$, $\Pi_1$, $\Pi_2\cdots$ stand for permutation patterns which are up to $N!$ cases.
    \item {\textbf{Output Layer}:}
    \begin{align}
    \label{output_equation}
    {\boldsymbol {\mathcal Z}}(t) = 
    {\boldsymbol {\mathcal G}}(t)\times_1{\boldsymbol W}_{out-1}\times_2 {\boldsymbol W}_{out-2}\cdots \times_N{\boldsymbol W}_{out-N}
    \end{align}
    \item {\textbf{Loss Function}:}
    \begin{align}
    \label{loss_general}
    \min_{\tau}\min_{{\boldsymbol W}_{out-1}, {\boldsymbol W}_{out-2}, \cdots,{\boldsymbol W}_{out-N}} \sum_{q=1}^{N_K}\sum_{t = 1}^{N_T}\|{\boldsymbol {\mathcal Z}}_q(t) - {\boldsymbol {\mathcal G}}_q(t+\tau)\times_1{\boldsymbol W}_{out-1}\cdots \times_N{\boldsymbol W}_{out-N}\|_F^2
    \end{align}
\end{itemize}

Similarly to the 2-mode case, the output weights are learned through alternating least squares. The optimization (\ref{loss_general}) can also be formulated as a high order tensor decomposition as defined in (\ref{loss_function}). Moreover, to further avoid model overfitting, regularization terms can be added in the loss function, such as ridge regression, i.e., $\|{\boldsymbol W}_{out-1}\|_F^2+ \|{\boldsymbol W}_{out-2}\|_F^2+\cdots+\|{\boldsymbol W}_{out-N}\|_F^2$. In addition, we can observe that the optimization problem (\ref{loss_general}) is not the canonical Tucker decomposition defined in \cite{de2008decompositions}. This is because the factor matrices are not designed as full column-rank in our framework. On the contrary, we choose the factor matrices with full row-rank to fulfill the mechanism of RC that is ``yielding desired output through dimension reduction from internal memory states.''

\subsection{Theoretic Analysis}

We now study the condition on the uniqueness of solving (\ref{loss_general}) via alternating least squares. Through our derivation as presented in Appendix, we can arrive at the following theorem.

\begin{theorem}
Given $\boldsymbol {\mathcal Z} \in {\mathbb C}^{N_{out-1} \times N_{out-2} \times \cdots \times N_{out-N}\times N_T \times N_K} $ and $\boldsymbol {\mathcal G} \in {\mathbb C}^{N_{f-1} \times N_{f-2} \times \cdots \times N_{f-N} \times N_T \times N_K}$ with rank-$(N_{out-1}, N_{out-2}, \cdots, N_{out-N},N_{T},N_{Q})$ \footnote{This stands for the multi-mode rank of a tensor. The definition can be found in Appendix.} and rank-$(N_{f-1}, N_{f-2}, \cdots, N_{f-N}, N_T, N_K)$ respectively, and $\forall n$, $N_{f-n}\geq N_{out-n}$, $N\geq2$, the achieved minimization of (\ref{loss_general}) is unique by using ALS when the initialization factor matrices are chosen as full rank and 
\begin{align}
\label{uniqueness_condition}
\sum_{i\neq n} N_{out-i} +N_T +N_K\geq N_{f-n}, \forall n.
\end{align} 
\end{theorem}

The above theorem reveals that the uniqueness condition of the Multi-Mode RC learning is characterized by the shape of the output tensor and feature core-tensor. Alternatively, if we use a single batch for training, i.e., merge the last two modes of the tensors into one, the shape of the output tensor and core-tensor respectively become $\boldsymbol {\mathcal Z} \in {\mathbb C}^{N_{out-1} \times N_{out-2} \times \cdots \times N_{out-N}\times (N_T N_K)} $ and $\boldsymbol {\mathcal G} \in {\mathbb C}^{N_{f-1} \times N_{f-2} \times \cdots \times N_{f-N} \times (N_T N_K)}$. Therefore, the uniqueness condition can be rewritten as \begin{align}
\label{uniqueness_condition2}
\sum_{i\neq n} N_{out-i} +N_TN_K\geq N_{f-n}, \forall n.
\end{align} 
In our experiments, we observe that when the uniqueness condition holds, the loss-value is often greatly larger than zero. The model thereby does not over-fit to the training dataset which can offer generalization on the unseen testing dataset. More related discussions on this observation are in the evaluation section. 

On the other hand, Multi-Mode RC can be analyzed  as an advance of conventional RC by imposing particular structures on the output layer, where the conventional RC refers to RC operating on single-mode data structures, i.e., scalars and vectors. To gain this insight, we consider vectorizing the output layer of a 2-mode RC. According to (\ref{output_mode2}), the resulting output equation of the 2-mode RC via a single-mode RC based formulation is given by,
\begin{align*}
    {\textit{Vec}}({\boldsymbol Z}(t)) = ({\boldsymbol W}_{out-2}\otimes{\boldsymbol W}_{out-1}){\textit{Vec}}({\boldsymbol G}(t)).
\end{align*}
The above equation reveals that the output weight of multi-mode RC is forged as a Kronecker product of two sub-matrices to process a ``vectorized'' ${\boldsymbol G}(t)$. As opposed to single-mode RC using a fully connected layer, the resulting Kronecker output layer is with less freedom on parameters which requires a less amount of data to fit. Meanwhile, the Kronecker structure can further reduce time and space complexity. We present the complexity analysis results in Table \ref{complex_table}, where we assume that the conventional RC and Multi-Mode RC are with the same input-output size. In this table, the time complexity of the forward path is calculated by the matrix product between the output layer and RC state at each sample, while the complexity on output learning is from the matrix inverse operations involved in solving the loss objectives of the entire training data set. The memory costs in the forward path are calculated based on the size of internal states and output weights. Meanwhile, the memory spent on learning is on the same scale as the size of the internal state. Moreover, the input buffer length is ignored in this table for simplicity. However, it can be easily calculated by substituting $N_{in-n}$ as $N_{in_n}T'$ in this table.
\begin{table}[h]
\centering
\caption{Time Complexity and Memory Usage Comparison }
\begin{tabular}{|c|c|c|}
\hline
NN Operations                    & Time                  & Memory                \\ \hline
Standard RC forward                          & ${\mathcal O}(\prod_n N_{out-n} (\prod_n N_{in-n}+N_s^N))$                     & ${\mathcal O}(\prod_n N_{out-n} (\prod_n N_{in-n}+N_s^N))$                     \\ \hline
Multi-Mode RC forward                        & ${\mathcal O}(\sum_n N_{out-n}(\prod_{n}N_{in-n}N!+N_s^N ))$                   & ${\mathcal O}(\sum_n N_{f-n}N_{out-n}+\prod_n N_{in-n}N!+N_s^N)$                     \\ \hline
Standard RC learning                         & ${\mathcal O}((\prod_n N_{in-n}+N_s^N)^3N_K)$& ${\mathcal O}((\prod_n N_{in-n}+N_s^N) N_K)$\\ \hline
\multicolumn{1}{|l|}{Multi-Mode RC learning} & ${\mathcal O}(\sum_n N_{f-n}^3\prod_nN_{out-n}N_K)$ & ${\mathcal O}(\prod_n N_{in-n}N!+N_s^N)$ \\ \hline
\end{tabular}
\label{complex_table}
\end{table}

\section{Application: Online Symbol Detection for Multi-user Massive MIMO}
\label{application_mimo_ofdm}
In this section, we will briefly review the transceiver architecture of multi-user massive MIMO-OFDM systems and elaborate on how to apply Multi-Mode RC to symbol detection of an \textbf{uplink} massive MIMO network.
\subsection{Multi-user Massive MIMO-OFDM System}
\begin{figure}
    \centering
    \includegraphics[width = 0.5\linewidth]{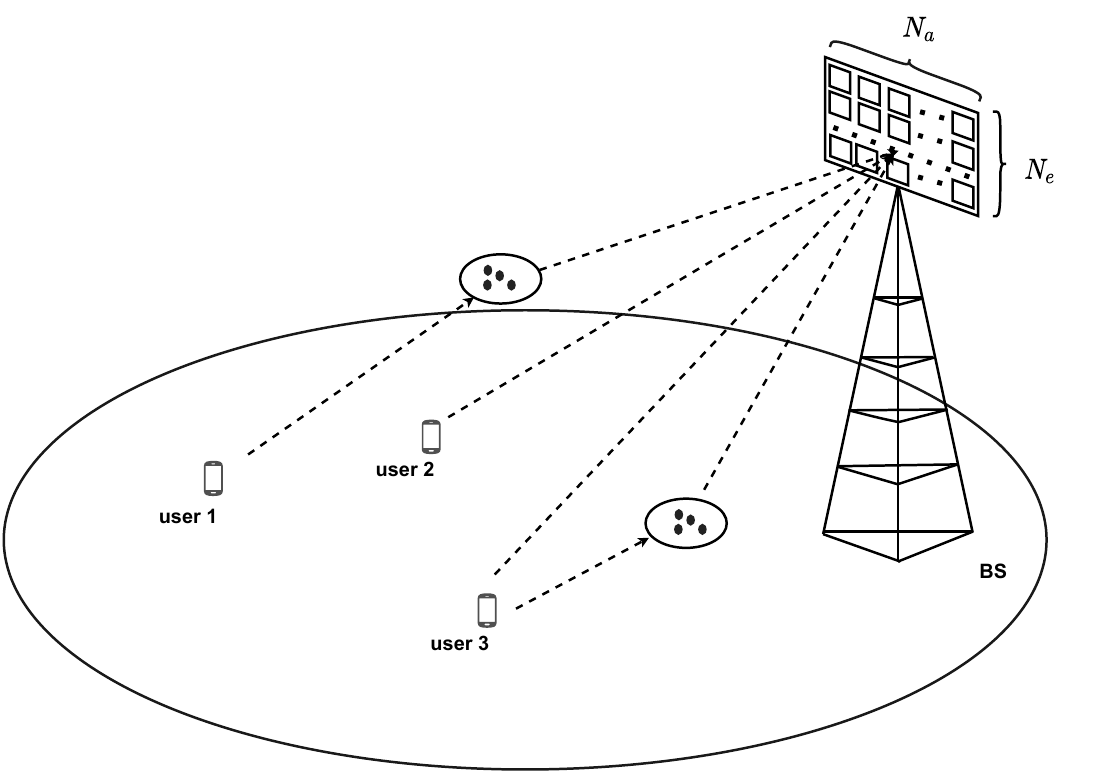}
    \caption{Uplink transmission in the multi-user massive MIMO system.}
    \label{multi_user_MIMO}
\end{figure}
\begin{figure}[t]
\centering
\includegraphics[width = 0.6\linewidth]{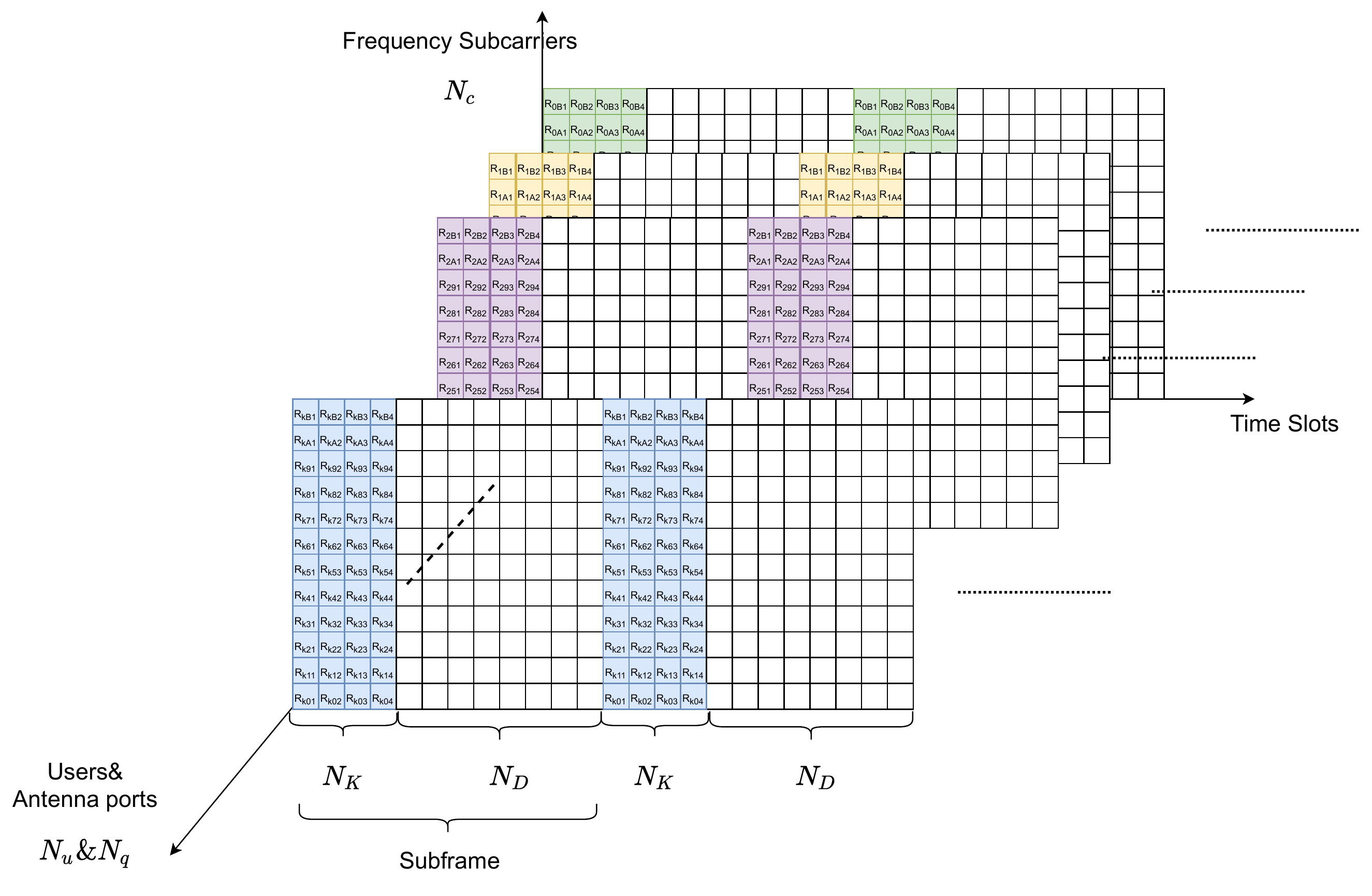}
\caption{Massive MIMO-OFDM resource grids (subframe-subcarrier) structure for RC training and symbol detection}
\label{dataset}
\end{figure}

We assume $N_u$ scheduled users are distributed in a cell communicating to a base station (BS) equipped with a massive rectangular array as shown in Fig. \ref{multi_user_MIMO}, where each user is mounted with $N_q$ antennas. The transmitted signals from all users to BS can be written as ${\boldsymbol X}(t) \in {\mathbb C}^{N_u \times N_q}$. 
Let ${\boldsymbol x}(t) = {\text{vec}}({\boldsymbol X}(t))\in {\mathbb C}^{N_t \times 1}$, where $N_t = N_u N_q$. 
Each entry of ${\boldsymbol x}(t)$ is a time sequence which stands for a stream of OFDM signals. For convenience, the OFDM signal ${\boldsymbol x}(t)$ is organized as OFDM resource grids as illustrated in Fig.~\ref{dataset}. 
In the OFDM resource grids, the horizontal direction represents OFDM symbols, while the vertical direction stands for sub-carriers indices. OFDM symbols are constructed into subframes where the time domain signal ${\boldsymbol x}(t)$ is obtained by applying an inverse Fourier transform (IFFT) on symbols across the subcarriers. Cyclic-prefix (CP) is added for each OFDM symbol. Let $N_{c}$ denote the number of subcarriers and $N_D + N_K$ be the number of OFDM symbols within a subframe.
Here, $N_K$ is the number of OFDM symbols that are used as pilots/reference signals whereas $N_D$ OFDM symbols within the same subframe are used for data transmission. Each element on the resource grids is modulated by quadrature amplitude modulation (QAM). Note that pilots/reference signals are used to conduct CSI estimation at the receiver in modern 4G and 5G networks.
Furthermore, \textbf{$N_K$ is configured to be smaller than $N_D$} to reduce over-the-air signaling overhead. 
Meanwhile, $N_K$ is often designed to be proportional to the number of streams to offer a reliable CSI estimation.

The received signal at the BS can be expressed as the following,
\begin{align}
\label{mimo_ofdm_signal}
{\boldsymbol {Y}}(t) = r \left(\sum_{\ell = 0}^L {\boldsymbol {\mathcal H}}(\ell)\times_{3}{\boldsymbol x}(t-\ell) + {\boldsymbol {\mathcal N}}(t)\right)
\end{align}
where $r(\cdot)$ is a function which characterizes the non-linearity at the receiver, such as ADCs, as well as model mismatch; ${\boldsymbol {\mathcal H}}(\ell) \in {\mathbb C}^{N_a \times N_e \times N_t}$ is a tensor which defines a spatial channel response at the $\ell$th-delay, where the total number of delays is denoted as $L$; $N_a$ is the number of azimuth antennas, $N_e$ is the number of elevation antennas of the massive MIMO BS antenna array.
Our objective is to train a NN $\mathcal D$ which can recover ${\boldsymbol X}(t)$ by using ${\boldsymbol Y}(t)$, i.e., 
\begin{align}
{\mathcal D}({\boldsymbol Y}(t)) = {\boldsymbol X}(t),
\end{align}
such that the NN is learned by
\begin{align}
\min_{\mathcal D} f\left(\left\{{\mathcal D}({\boldsymbol Y}(t))\}, \{{\boldsymbol X}(t)\right\}\right).
\end{align}

\subsection{Online Symbol Detection by Multi-Mode RC}
Multi-Mode RC serves as the detection NN, $\mathcal D$, through the over-the-air pilots/reference signals on a subframe basis. The pilots defined in existing massive MIMO-OFDM systems of each subframe is directly utilized as the training dataset. The following data symbols in the same subframe is the testing dataset, i.e., we train the NN using $N_K$ pilots to detect $N_D$ data symbols in each subframe. This constraint makes the learning framework different from conventional NNs to enable online learning for robust and adaptive communications. The azimuth direction is set as the first mode of Multi-Mode RC and the elevation direction is set as the second mode. Each OFDM pilot symbol is considered as one training batch. Accordingly, the input sequence length equals the number of subcarriers, $N_c$, plus CP. The output is then truncated to be a $N_c$-length sequence following the process as described under equation (\ref{loss_2_mode}). The symbols of each stream on each subcarrier are obtained through quantization and demodulation. 

In massive MIMO systems, symbol detection can be conducted through either a joint or a decomposed approach:
\subsubsection{Joint Processing}
In the joint processing, data symbols are obtained through a single Multi-Mode RC with a multi-head output, where the size of the first and second mode of the RC output sequence are respectively $N_q$ and $N_u$. The mode order of the output node also can be reversely configured. This is because Multi-Mode RC treats equally on each output mode according to the generation rule of the internal feature queue as shown in (\ref{feature_queue}). 
A well trained joint model is anticipated to yield a good symbol detection performance since all interference and imperfect factors are handled jointly. 
\subsubsection{Decomposed Processing} The decomposed approach refers to learning the output weight through a decomposed way. For instance, in the case of 2-mode RC, it has $N_u \times N_q$ pairs of $({\boldsymbol w}_{out-1}, {\boldsymbol w}_{out-2})$ to learn, where the vector weight ${\boldsymbol w}$ maps the internal states to a scalar entry of the output tensor sequence. In this framework, the training on each decomposed output weight is based on their individual loss allowing the training through a parallel manner which can significantly reduce the computation latency. On the other hand, the decomposed method takes extra resources on storage and computation compared to the joint approach. For instance, in 2-mode RC based joint approach, the size of output weight matrices ${\boldsymbol W}_{out-1}$ and ${\boldsymbol W}_{out-2}$ are respectively $N_{out-1} \times N_{f-1}$ and $N_{out-2} \times N_{f-2}$. In the decomposed way, the shapes of output mapping on each mode are respectively $1\times N_{f-1}$ and $1\times N_{f-2}$. Thus, there are $N_{out1}N_{out2} \times (N_{f-1}+N_{f-2})$ output weights in total for the decomposed approach which is higher than $N_{out-1} \times N_{f-1} + N_{out-2} \times N_{f-2}$ from the joint way. Overall, the map from MIMO-OFDM parameters to the Multi-Mode RC parameters is summarized in Table \ref{parameter_map}.

\begin{table}[] 
\centering
\caption{Notations of Multi-Mode RC based MIMO-OFDM Symbol Detection}
\begin{tabular}{|l|l|l|}
\hline
Notations & {Definition} & Corresponding notations in Multi-Mode RC \\ \hline
$N_t$ & Number of antennas stacked from all users & N/A\\\hline
$N_u$ & Number of users & $N_{out-1}$ in joint processing\\\hline
$N_q$ & Number of antennas at each user & $N_{out-2}$ in joint processing\\\hline
$N_a$ & Number of azimuth antennas & $N_{in1}$\\\hline
$N_e$ & Number of elevation antennas & $N_{in2}$\\\hline
$N_c$ & Number of OFDM sub-carriers & $N_T$\\\hline
$N_K$ & Number of pilot symbols in a frame & $N_K$-Training batches (if use multi-batch training)\\\hline
$N_D$ & Number of data symbols in a frame & Testing batches \\ \hline
${\boldsymbol x}(t) \in {\mathbb C}^{N_t\times 1}$ & Transmitted Signal & Desired output\\\hline
${\boldsymbol X}(t) \in {\mathbb C}^{N_u\times N_s}$ & Transmitted Signal & Desired output \\\hline
${\boldsymbol Y}(t) \in {\mathbb C}^{N_a\times N_e}$ & Received Signal & Input\\ \hline
${{\boldsymbol {\mathcal Y}}} \in {\mathbb C}^{N_a\times N_e \times T}$  &  Stacked tensor of received signal& Input \\  \hline
\end{tabular}
\label{parameter_map}
\end{table}

\section{Performance Evaluations}
\label{Evaluations}
This section provides performance evaluations of the introduced Multi-Mode RC for uplink symbol detection in a multi-user massive MIMO-OFDM scenario. 
We choose uncoded bit error rate (BER) as the quantitative metric to evaluate the reliability of the underlying link. Table \ref{parameter_map} contains simulation parameters of the massive MIMO-OFDM system. The default system configuration is: $N_a = 8$, $N_e = 8$, $N_{c} = 512$, $N_{cp} = 32$, $N_q = 2$, $N_u = 2$, $N_K = 4$ and $N_D = 12$. Note that in this setting, the pilots/reference signals overhead is $25\%$ which is inline with 5G standards~\cite{5GNR2}. The channel coefficients are generated according to the clustered delay line (CDL) model defined in 3GPP Technical Report (TR) 38.901, where the transmitter and receiver are configured with uniform linear arrays having half-wavelength antenna spacing, and the power delay profile is configured with a cluster delay rate of $3$. The maximum delay spread of the channel is set as the length of the CP in OFDM. Each obtained BER point is collected over $100$ consecutive subframes. SNR is defined as the average power ratio between noise-free received signal and the additive noise. The configuration of the Multi-Mode RC is set as follows: $T' = N_{CP}$, $N_s = 8$, the number of ALS iterations is set as $6$, the state transition matrix ${\boldsymbol W}_{tran}$ on each mode is independently generated with its spectral radius less than $1$. Meanwhile, the input weights matrix ${\boldsymbol W}_{in}$ is generated independently for each mode from a uniform distribution on $[-1, 1]$. 

\subsection{Parsing Multi-Mode RC}
\begin{figure}[h]
    \centering
    \includegraphics[width=0.7\linewidth]{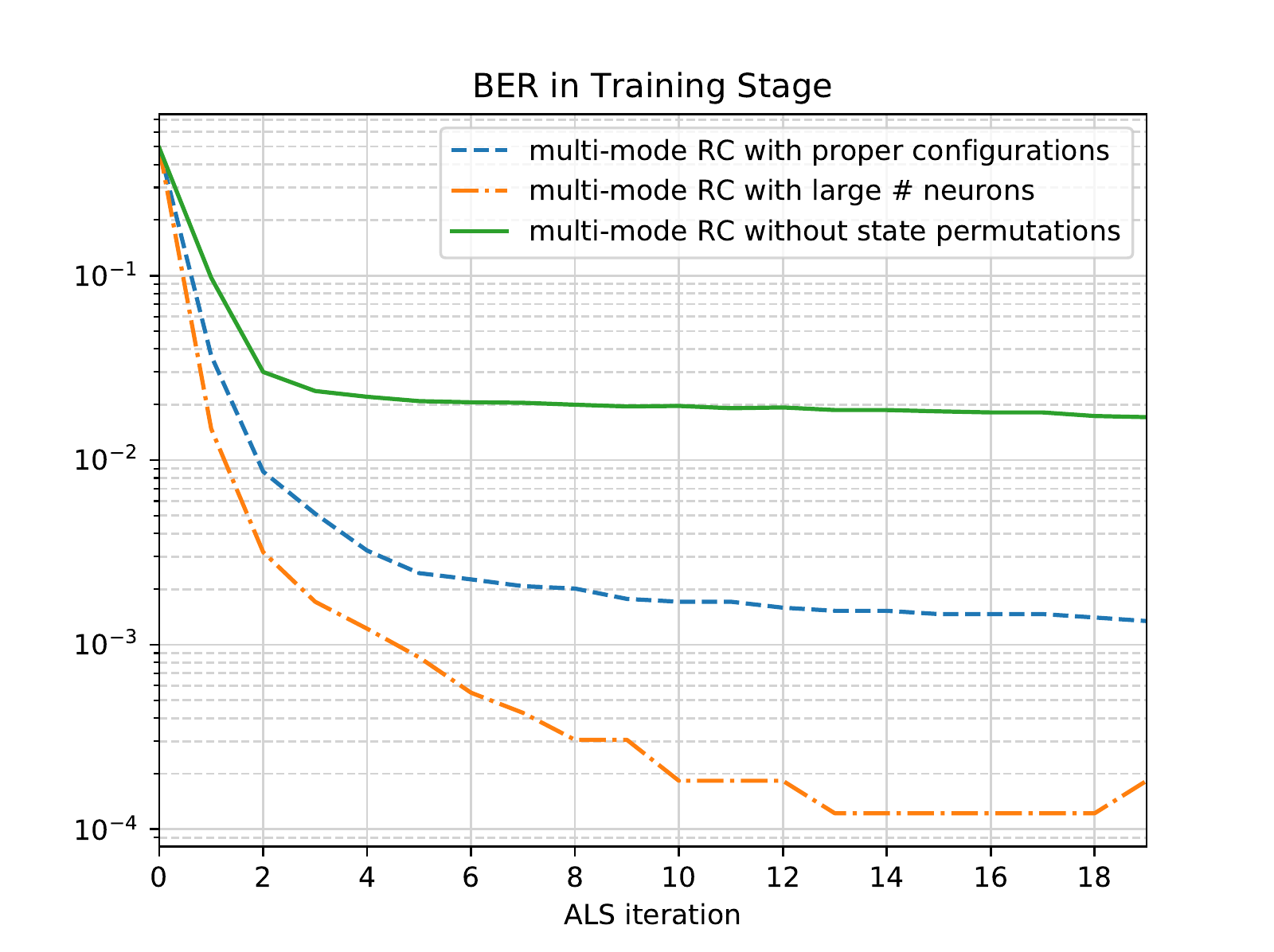}
    \caption{Training BER of multi-mode RC with respect to iterations in ALS.}
    \label{training_ber}
\end{figure}

\begin{figure}[h]
    \centering
    \includegraphics[width=0.7\linewidth]{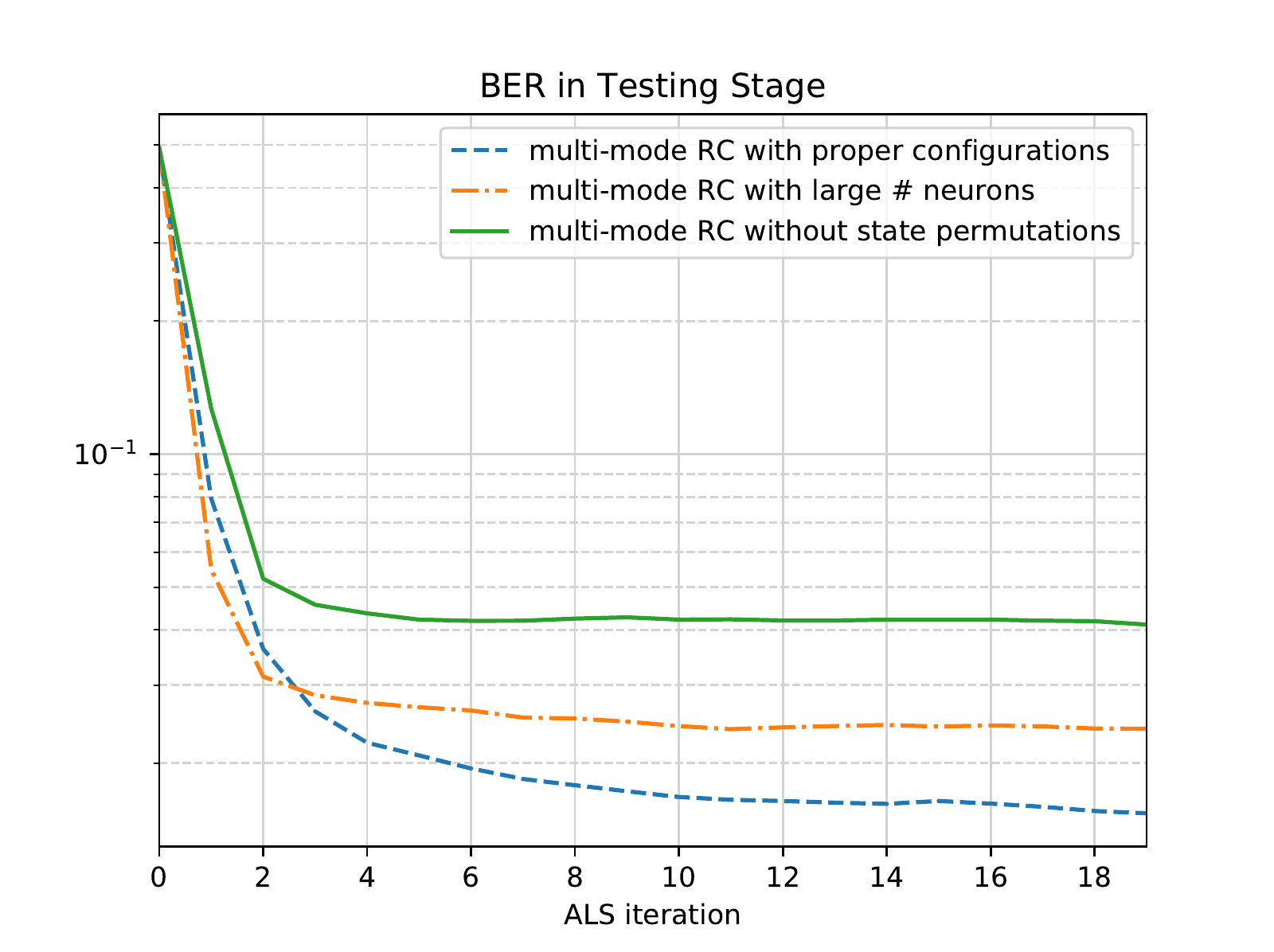}
    \caption{Testing BER of Multi-Mode RC with respect to iterations in ALS.}
    \label{testing_ber}
\end{figure}

We first parse various components of a Multi-Mode RC to offer more insights on the underlying NN structure. We consider three types of Multi-Mode RC in the evaluation: 1) Our introduced one; 2) Our introduced one without using the tensor permutation to construct the feature queue in (\ref{feature_queue}); 3) Our introduced one using a large number of $N_s$, where $N_s = 128$. Fig.~\ref{training_ber} and Fig.~\ref{testing_ber} respectively show the training and testing BERs under different numbers of iteration when $SNR = 15{\text {dB}}$. As we can observe that without tensor permutation in the feature queue, the RC performs underfitting to the task. This is because the feature from different modes of the input tensor sequence has not been equally extracted. Meanwhile, if we increase the number of neurons, the NN model complexity increases. Accordingly, it ``overfits'' the training data as the training BER dramatically decreases, whereas the testing BER increases.

\subsection{Uniqueness Conditions}
\begin{figure}[h]
    \centering
    \includegraphics[width = 0.7\linewidth]{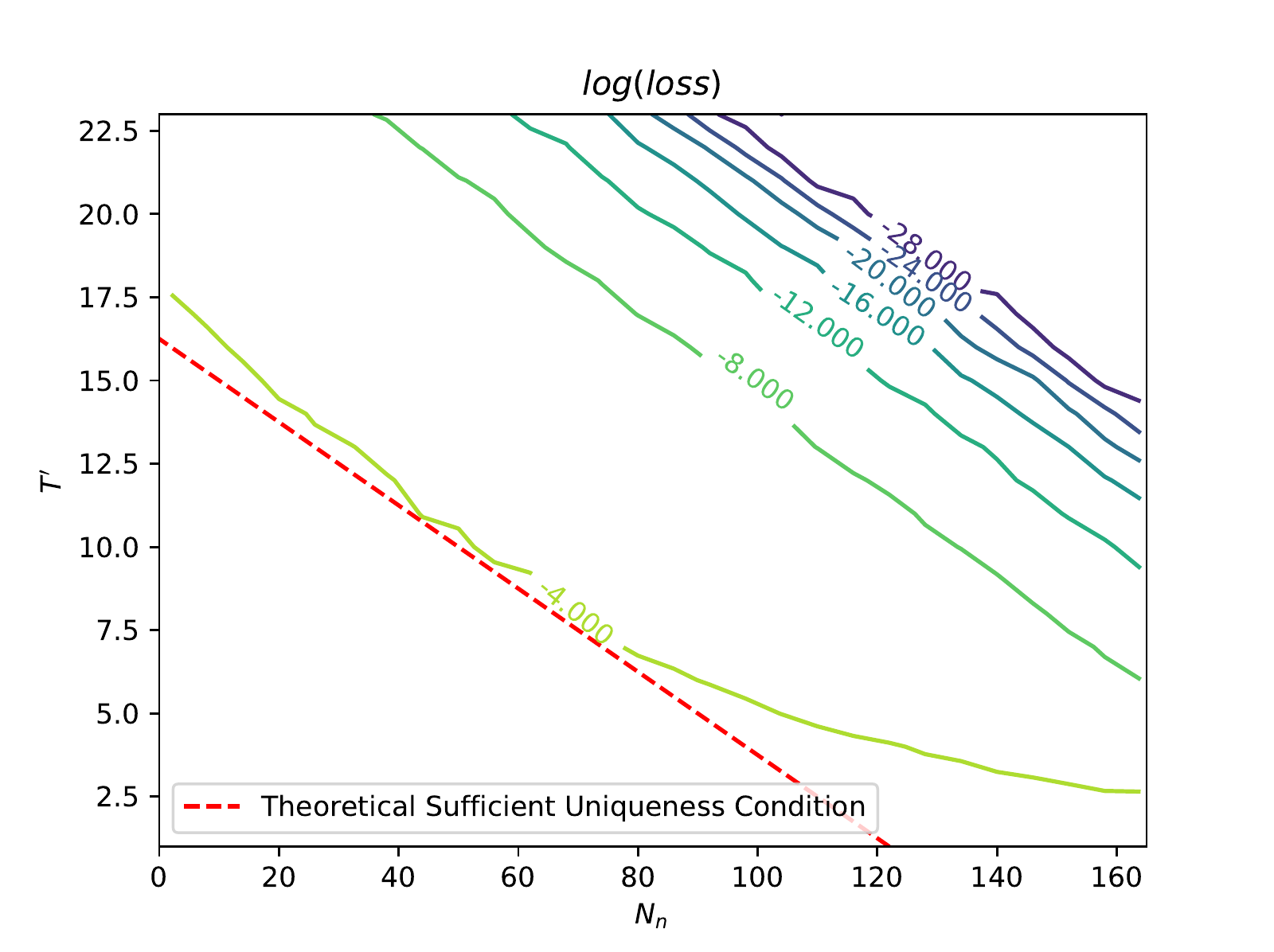}
    \caption{Log loss contour in $(N_s, T')$ plane in training stage}
    \label{ber_contour1}
\end{figure}

\begin{figure}[h]
    \centering
    \includegraphics[width = 0.7\linewidth]{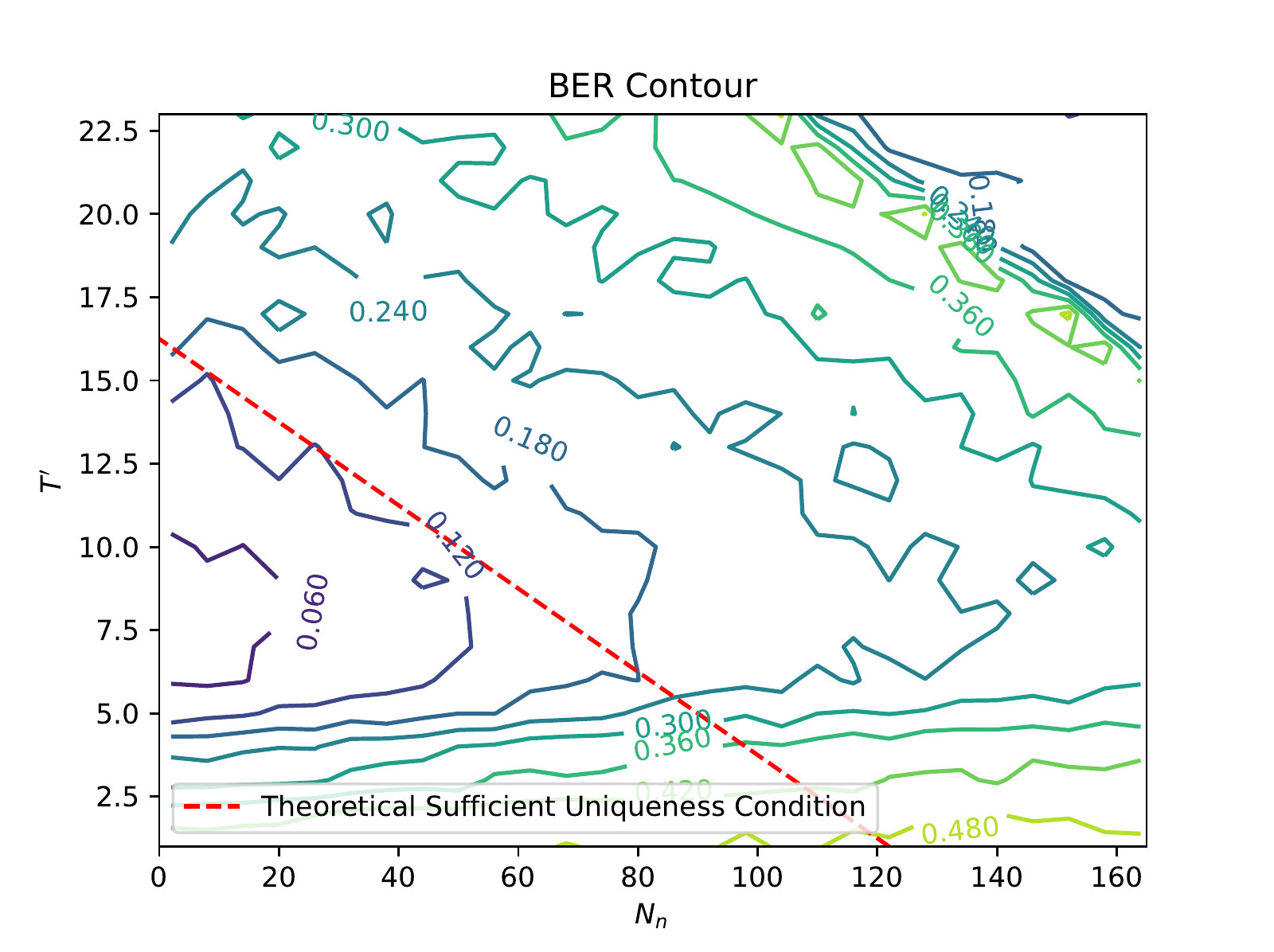}
    \caption{BER contour in $(N_s, T')$ plane in testing stage}
    \label{ber_contour2}
\end{figure}

Now, we investigate how the uniqueness condition defined in Theorem 1 determines the training and testing performance. 
For convenience, we set $N_c = 64$ and use single-batch based training in this evaluation. 
According to (\ref{uniqueness_condition2}), the critical conditions for this task becomes, 
\begin{align*}
N_u + N_TN_K == 2N_s + N_aT' + N_eT'\\
N_q + N_TN_K == 2N_s + N_aT' + N_eT'
\end{align*}
When we only change parameters $T'$ and $N_s$ while fixing the rest based on our default setup, the critical conditions become $130 == 8T'+ N_s$. 
This condition is plotted as the dashed red line in Fig.~\ref{ber_contour1} and Fig.~\ref{ber_contour2}.
Meanwhile, we also plot the contours of the log-loss as well as the BER in the same $(N_s,T')$ plane. 
As shown in Fig.~\ref{ber_contour1}, the loss is guaranteed to be greater than a threshold, e.g., $-4.00$, when the uniqueness condition holds. 
On the other hand, when the condition is violated, the loss tends to be close to zero. 
In this case, the RC model overly fits to the training data which brings a high risk of over-fitting. 
This result is also consistent with the BER contour plotted in Fig.~\ref{ber_contour1}. 
Note that even though satisfying the uniqueness condition can potentially avoid overfitting, it may cause underfitting as we can observe the high BER below the condition line in Fig.~\ref{ber_contour2}.

\subsection{Comparison with State-of-Art Detection Strategies}
\begin{figure}
\centering
\includegraphics[width=0.7\linewidth]{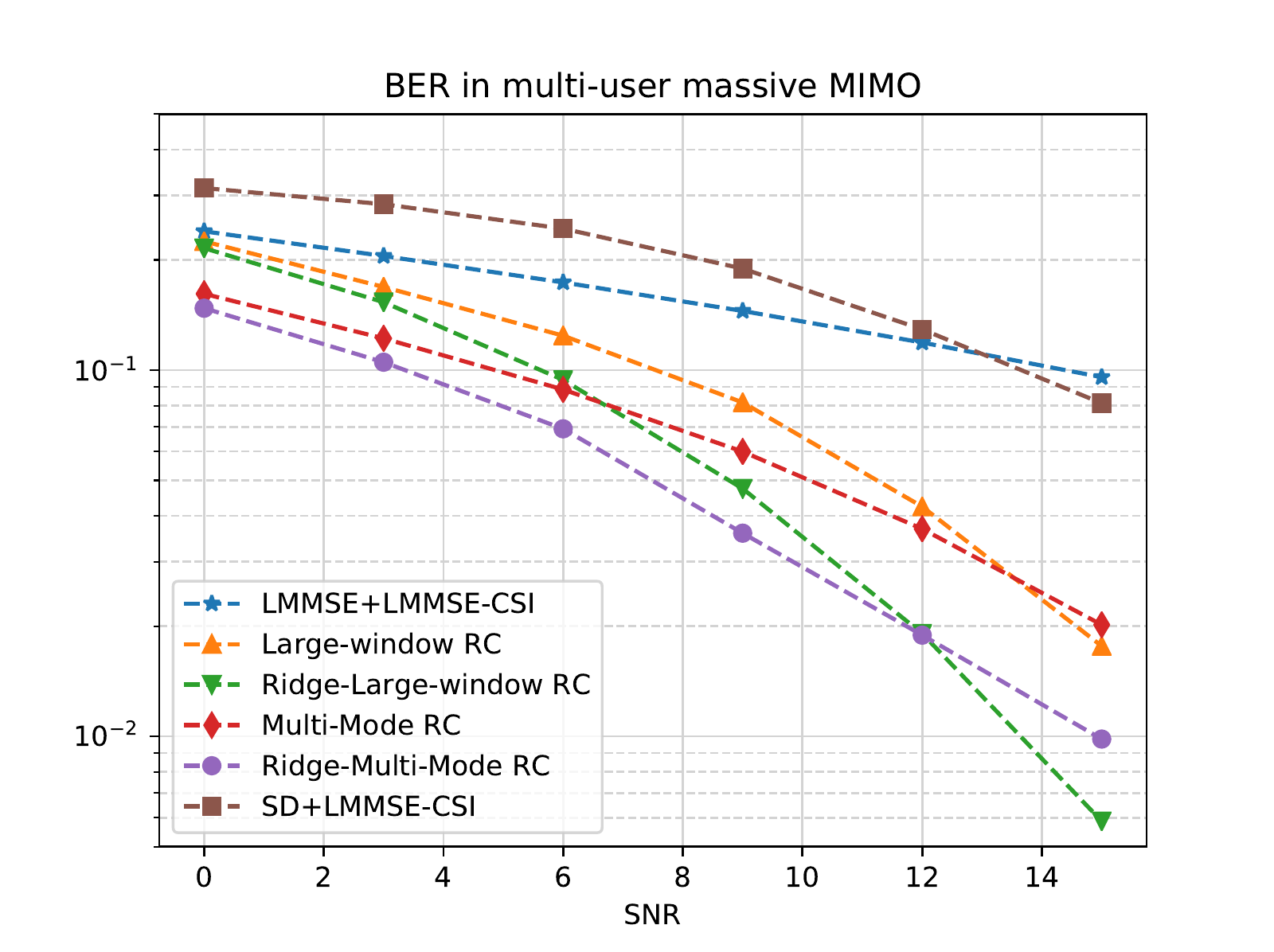}
\caption{BER in a multi-user massive MIMO system with 64 antennas at the BS ($8 \times 8$ antenna array).}
\label{mu_mimo_snr}
\end{figure}
We now investigate the BER versus SNR using different approaches. 
In Fig.~\ref{mu_mimo_snr}, the compared methods are:
1) LMMSE+LMMSE-CSI which uses linear minimum mean square error (LMMSE) based symbol detection under the LMMSE estimated CSI. 2) SD+LMMSE-CSI which uses sphere decoding for symbol detection based on the LMMSE estimated CSI. 3) Large-window RC refers to the windowed echo state network (WESN) introduced in \cite{zhou2019} by vectorizing the input as a vector and setting the input buffer size as $52$. 4) Ridge-large-window RC refers to the same WESN but using $l_2$ norm as a penalty term to the output weights in the loss function. 5) Multi-Mode RC is the introduced method. 6) Ridge-Multi-Mode RC standards for the same Multi-Mode RC but also adding a $l_2$ norm as the regularization on the output weights in the loss objective. 
Fig.~\ref{mu_mimo_snr} clearly demonstrates the performance gain of the Multi-Mode RC over the signal processing-based methods (LMMSE+LMMSE-CSI and SD+LMMSE-CSI). 
Meanwhile, we can see that the Multi-Mode RC is more robust than the single-mode RC as the multi-mode feature of MIMO-OFDM signals is leveraged for the symbol detection. 

\begin{figure}
\centering
\includegraphics[width=0.7\linewidth]{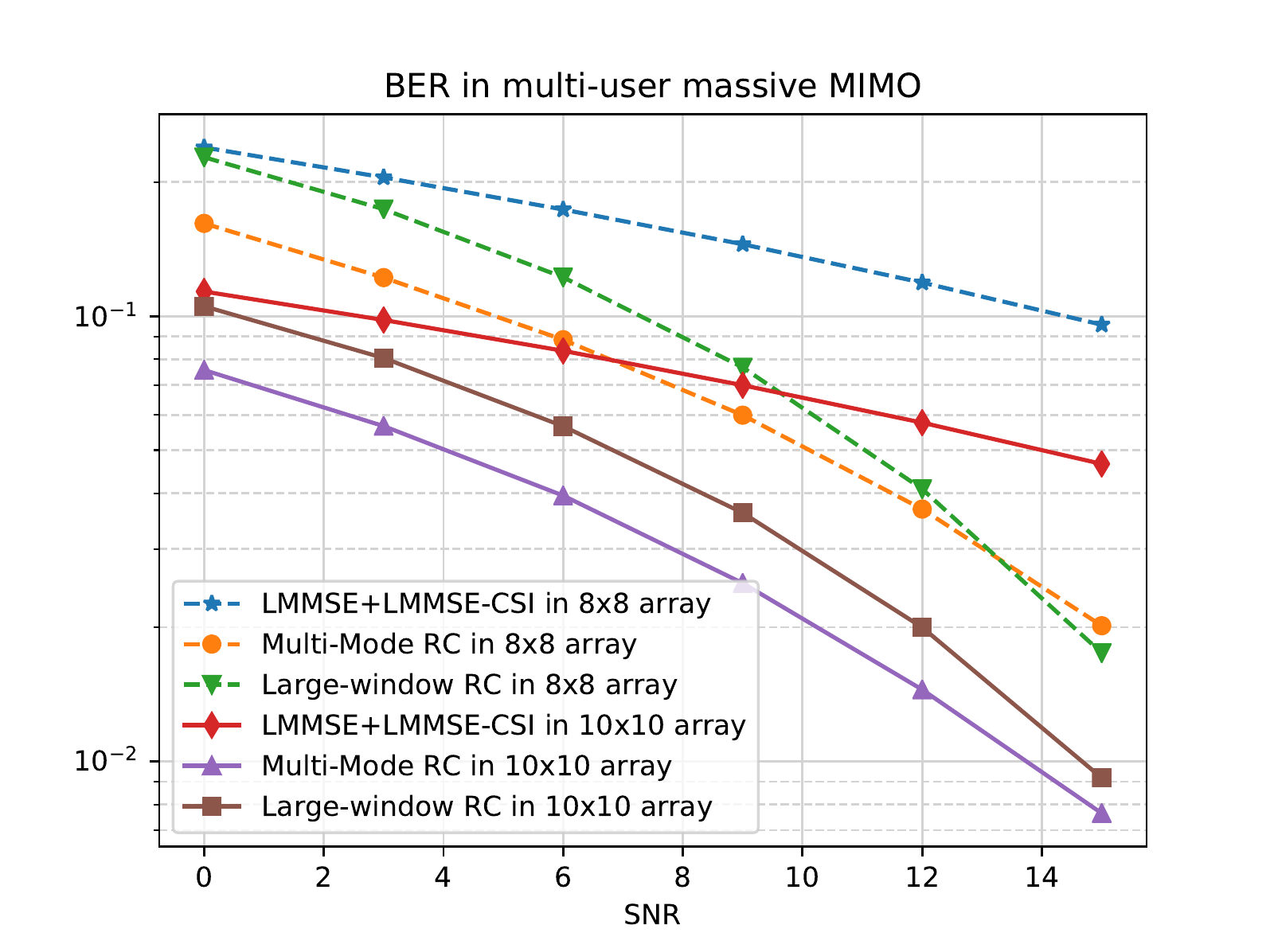}
\caption{BER in multi-user massive MIMO with $64$ ($8 \times 8$) and $100$ ($10 \times 10$) antennas at the BS.}
\label{mu_mimo_snr2}
\end{figure}
In addition, we investigate the BER performance by increasing the array size. 
Fig. \ref{mu_mimo_snr2} shows that when the number of antennas increases (e.g improve the antenna array from $8\times 8$ to $10 \times 10$), the BER curves of all methods are improved. 
On the other hand, the Multi-Mode RC continues showing its advantage over other methods. 
 
\begin{figure}[t]
\centering
\includegraphics[width=0.7\linewidth]{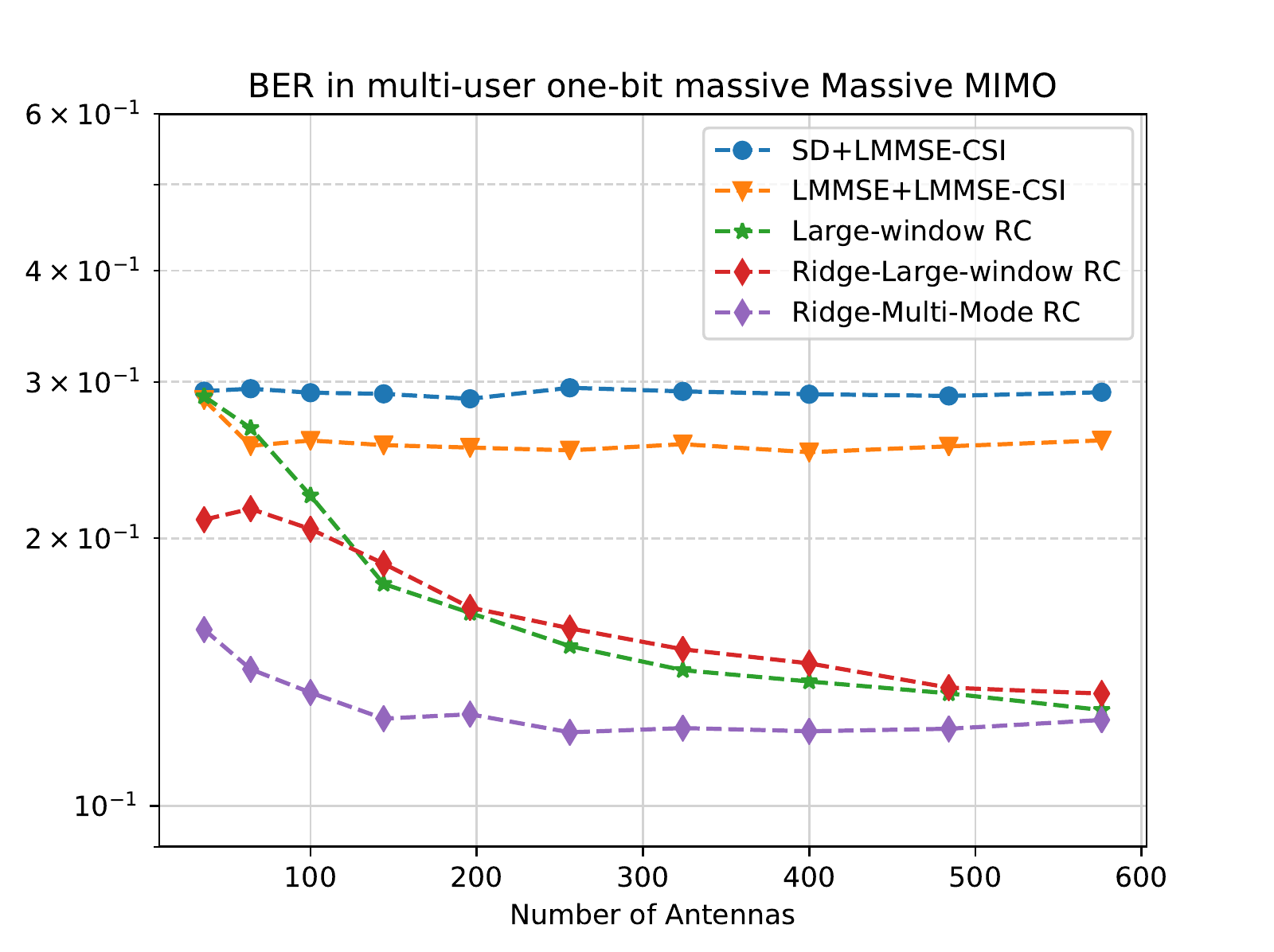}
\caption{BER in one-bit multi-user massive MIMO systems with different antenna numbers.}
\label{one_bit_massive_mimo}
\end{figure}

\subsection{Performance Evaluation under Receiving Non-linearity ---- Low Resolution ADCs}
To show the advantage of RC-based approach in other model mismatch scenarios, we evaluate the BER performance when low precision quantization is added in the link which is extremely relevant to massive MIMO systems. 
We consider the extreme case of using one-bit ADC which quantizes the in-phase and quadrature components to 1 or -1. 
The definition of the quantizer for any one of the components is,
\begin{equation}
q(x) = A_{max}\cdot {\text{sign}}(x)
\end{equation}
where $A_{max}$ is the maximum magnitude of the quantizer where we set it as $0.6$ in the evaluation. 
Fig.~\ref{one_bit_massive_mimo} clearly shows that the Multi-Mode RC is the most robust method in this scenario. 
Meanwhile, we can observe the saturation phenomena of the BER curve when the antenna number increases. 
This is because the quantization level on each antenna is set as a fixed value. 
Intuitively, the performance can be further improved by optimizing the quantization level on each antenna. 
This will be considered as our future work.  

\subsection{Comparison with Model-based Learning Approaches}
\begin{figure}
\centering
\includegraphics[width=0.7\linewidth]{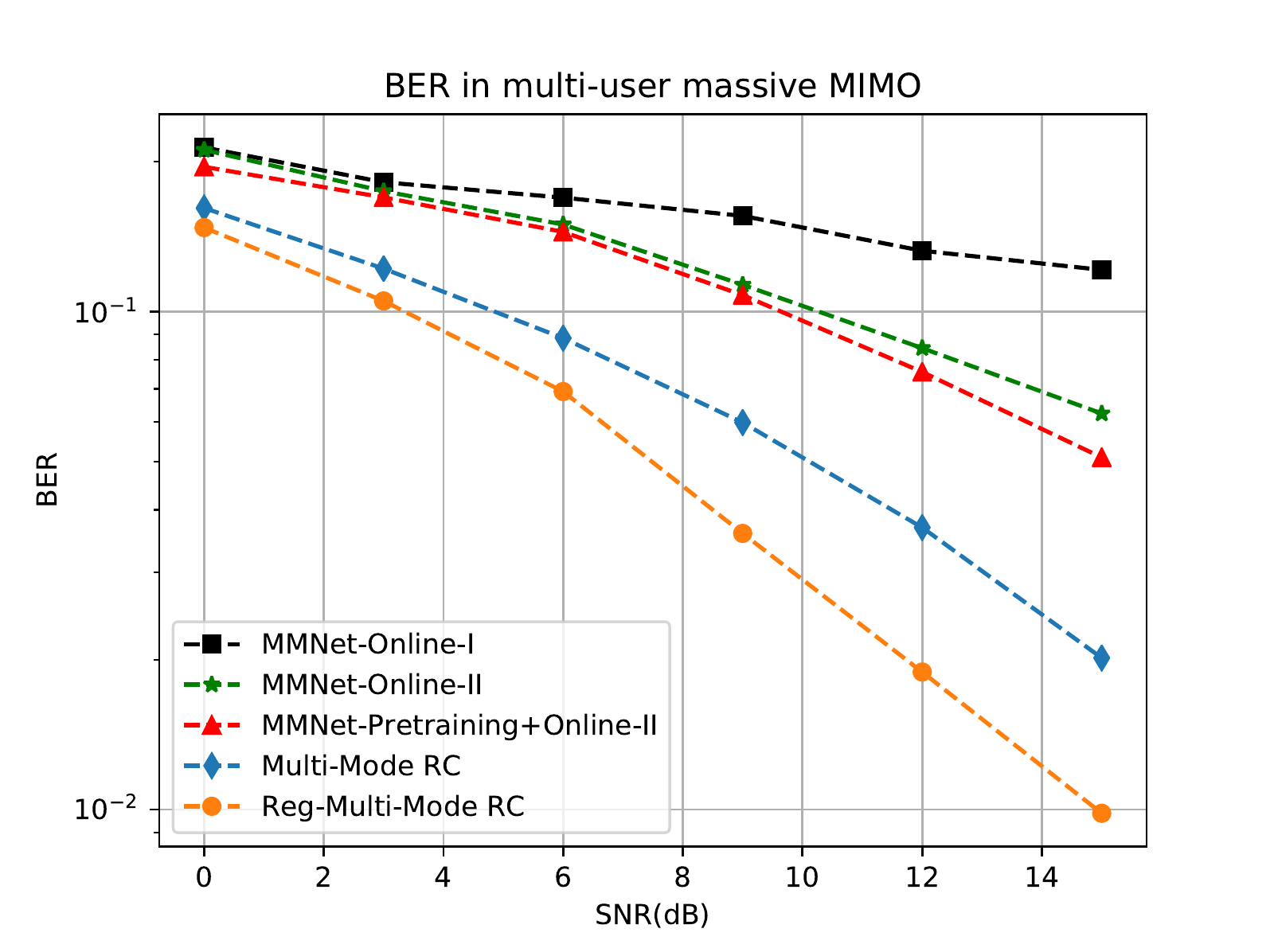}
\caption{BER in a multi-user massive MIMO system with 64 antennas at the BS ($8 \times 8$ antenna array).}
\label{mmnet}
\end{figure}

In this section, we compare the Multi-Mode RC against a state-of-the-art model-based symbol detection NN, MMNet \cite{khani2019adaptive}. 
MMNet is a deep NN structure based on unfolding iterative soft-thresholding algorithms, which adds degrees of flexibility on certain parameters in the NN for training. 
In our evaluation, MMNet is configured using the aforementioned training dataset associated with the LMMSE-estimated CSI. 
As the legends shown in Fig.~\ref{mmnet}, we choose three MMNet operation modes as the benchmark methods:
1) MMNet-Online-{I} contains only scalar trainable parameters. It assumes the channel additive noise in both testing and training stages are with homogeneous distributions. The ``Online'' scheme refers to a training framework described in the paper, by which a NN for conducting symbol detection on the first subcarrier is trained from scratch using 1000 iterations and $N_K$ pilot symbols, while other NNs with respect to the remaining sub-carriers are fine-tuned based on the first NN with 3 additional iterations using their individual training symbols. 2) MMNet-Online-{II} has matrix-form parameters as noise variance estimators per layer, which is considered as an advanced structure of MMNet-Online-{I} using the same training strategy. 3) MMNet-Pretraining-Online-{II} adds a pre-trained training stage to MMNet-Online {II}. In this way, the NNs are initialized with pre-trained weights using 256 pilot symbols from historical training datasets with different channel realizations.

Note that, while MMNet achieves notable performance on both i.i.d Gaussian channels and spatially-correlated channels in \cite{khani2019adaptive}, the utilized training symbols are relatively larger than the setup in this paper (e.g. $500$ pilot symbols associated with perfect CSIs which is not consistent to the online over-the-air scenario as in our paper). Furthermore, the computation and memory requirements on MMNet are significantly higher than the introduced method. To be specific, MMNet requires $N_c$ NNs to estimate symbols on all subcarriers, each of which stacks $10$ layers of neurons. Therefore, the number of training iterations significantly increases. In our method, we only use $4$ pilot symbols and a single NN to jointly accomplish the symbol detection task. Our evaluation results demonstrate that the Multi-Mode RC outperforms MMNet in the multi-user massive MIMO scenario with a steep performance improvement slope. Overall, Reg-Multi-Mode RC shows $5-6$ dB gain in SNR compared with MMNet.

\section{Conclusion}
\label{Conclusion}
In this paper, we presented a NN structure, Multi-Mode RC, for symbol detection in massive MIMO-OFDM systems. We elaborated on the NN architecture and its configuration for the symbol detection task. The introduced Multi-Mode RC framework is shown to be able to effectively cope with the model mismatch, waveform distortion as well as interference in the systems. Numerical results demonstrated the advantages of Multi-Mode RC in the following aspects: It can offer lower BER than conventional single-mode RC frameworks in low SNR regime while achieving reduced computational complexity. Compared to other model-based learning approaches, the introduced method can operate on a subframe-basis thus completely relying on the limited over-the-air pilots/reference symbols.
This attractive feature enables us to train the symbol detection task using a compatible signal overhead as modern cellular networks.

In our future work, we will consider the optimization of the quantization thresholds at each antenna port. Since quantization is an irreversible process, adaptive quantization strategies are a promising approach to preserve the waveform information. 
Furthermore, incorporating gradient-free learning algorithms into the RC framework is another interesting direction. 


%


\appendix[Proof of Theorem 1]
\begin{lemma}
\label{lemma1}
Given two full-rank matrices ${\boldsymbol X}\in {\mathbb C}^{M_1 \times M_2}$ and ${\boldsymbol G}\times {\mathbb C}^{H \times M_2}$, where $H\geq M_1$, the following least squares problem has a unique solution if and only if $rank({\boldsymbol X}) = rank({\boldsymbol W}^{\star})$, where ${\boldsymbol W}^{\star}$ is the optimum. 
\begin{align}
    \label{classic_ls}
    \min_{\boldsymbol W}\|\boldsymbol X - {\boldsymbol W}{\boldsymbol G}\|_F
\end{align}
\end{lemma}
\begin{IEEEproof}
Based on the assumptions, we have $rank({\boldsymbol X}) = \min \{M_1, M_2\}$ and $rank({\boldsymbol G}) = \min\{H, M_2\}$. In general, (\ref{classic_ls}) has a unique solution if and only if $rank({\boldsymbol G}) = H$ \cite{meyer2000matrix}.

$rank({\boldsymbol G}) = H$ implies $M_2\geq H$. Accordingly, the solution is given by ${\boldsymbol X}{\boldsymbol G}^{+}$, where ${\boldsymbol G}^{+}$ represents the Moor-Penrose inverse of matrix $\boldsymbol G$. Since $\boldsymbol X$ is with full row rank, ${\boldsymbol G}^{+}$ is with full column rank and $ H \geq M_1$, we have $rank({\boldsymbol X}{\boldsymbol G}^{+}) = rank({\boldsymbol X})$. Thus, we have $rank({\boldsymbol W}^{\star}) = rank({\boldsymbol X}{\boldsymbol G}^{+}) = rank({\boldsymbol X})$.

On the contrary, suppose $rank({\boldsymbol W}^{\star}) = rank({\boldsymbol X})$ holds. We have $H\leq M_2$, otherwise there exists ${\boldsymbol W}^{\star}+{\boldsymbol H}$ which is also a minimum of (\ref{classic_ls}), where $\boldsymbol H$ is a non-zero solution of ${\boldsymbol H}{\boldsymbol G} = {\boldsymbol 0}$. Since $\boldsymbol G$ is assumed with full rank, we have $rank({\boldsymbol G}) = H$. 
\end{IEEEproof}

\begin{lemma}
Given the same assumption as Lemma \ref{lemma1}, the sufficient and necessary condition for the uniqueness of (\ref{classic_ls}) can be expressed as 
\begin{align}
    M_2\geq H
\end{align}
\end{lemma}
\begin{IEEEproof}
Since the necessary and sufficient condition for the uniqueness is $rank({\boldsymbol G}) = H$. Therefore, the uniqueness condition can be alternatively written as (\ref{classic_ls}) is $M_2 \geq H$. 

\end{IEEEproof}
Before proceeding on the proof of Theorem 1, we introduce the following concepts to characterize rank properties of tensor \cite{de2008decompositions}.
\begin{definition}
The mode-n rank of a tensor $\boldsymbol{\mathcal G}$ is the mode-n unfolding of $\boldsymbol G$, i.e., ${\boldsymbol G}_{(n)}$.
\end{definition}
\begin{definition}
A N-order tensor $\boldsymbol{\mathcal G}$ is with rank-$(M_1,M_2, \cdots, M_N)$ when its mode-1 rank, mode-2 rank to mode-N rank are equal to $M_1$, $M_2$, and $M_N$, respectively.
\end{definition}
\begin{lemma}
Given $\boldsymbol {\mathcal X} \in {\mathbb C}^{M_1 \times M_2 \times \cdots \times M_N}$ and $\boldsymbol {\mathcal G} \in {\mathbb C}^{H_1 \times H_2 \times \cdots \times H_N}$, which are with rank-$(M_1, M_2, \cdots, M_N)$ and rank-$(H_1, H_2, \cdots, H_N)$ respective, and $H_n\geq M_n$, $N>2$, the minimization of 
\begin{align}
\label{tucker_opt}
\min_{{\boldsymbol W}_1, \cdots,{\boldsymbol W}_N} \|\boldsymbol{\mathcal X} - \boldsymbol{\mathcal G} \times_{1} \mathbf{W}_1 \times_{2} \mathbf{W}_2 \cdots \times_{N} \mathbf{W}_N\|_F
\end{align}
is unique if 
\begin{align}
    rank({\boldsymbol X}_{(n)}) = rank({\boldsymbol W}_n^{\star})
\end{align}
where ${\boldsymbol W}_1^{\star}$, $\cdots$, ${\boldsymbol W}_N^{\star}$ is the optimum. 
\end{lemma}
\begin{IEEEproof}
We prove this theorem by mathematical induction. According to Lemma 1, the uniqueness holds for $\min_{{\boldsymbol W}_1}\|{\boldsymbol { \mathcal X}} - {\boldsymbol { \mathcal G}}\times_1{\boldsymbol W}_1\|_F$. Then, we assume it holds for $N$ order tensor $\boldsymbol {\mathcal X}$ and $\boldsymbol {\mathcal G}$ with $N-1$ factor matrices. 

Now, we  consider the case with $N$ factor matrices,
\begin{align*}
    &\min_{{\boldsymbol W}_1, \cdots,{\boldsymbol W}_N} \|\boldsymbol{\mathcal X} - \boldsymbol{\mathcal G} \times_{1} \mathbf{W}_1 \times_{2} \mathbf{W}_2 \cdots \times_{N} \mathbf{W}_N\|_F \\
    &=\min_{{\boldsymbol W}_N} \{\min_{{\boldsymbol W}_{1},\cdots, {\boldsymbol W}_{N-1}}\|\boldsymbol{X}_{(N)} - \mathbf{W}_N  {\boldsymbol G}_{(N)} (\mathbf{W}_1 \otimes \cdots \otimes  \mathbf{W}_{N-1})^T\|_F\}
\end{align*}
We denote ${\boldsymbol W}_N^{\star}$ as one of the optima of the above problem. Since ${\boldsymbol W}_N^{\star}$ is with full row rank based on the assumption, we have tensor $\boldsymbol {\mathcal G}\times_N {\boldsymbol W}_N^{\star}$ with rank-$(H_1, H_2,\cdots,M_N)$. This is because ${\boldsymbol G}_{(n)}$ is with full row rank. Therefore, $({\boldsymbol W}_1^{\star}$, $\cdots$, ${\boldsymbol W}_{N-1}^{\star})$ as an optimum of $\min_{{\boldsymbol W}_1, \cdots,{\boldsymbol W}_{N-1}} \|\boldsymbol{\mathcal X} - \boldsymbol{\mathcal G} \times_{1} \mathbf{W}_1 \times_{2} \mathbf{W}_2 \cdots \times_{N} \mathbf{W}_N^{\star}\|_F$ is unique based on the inductive assumption. We then assert ${\boldsymbol W}_N^{\star}$ is unique. Otherwise it contracts to Lemma 1. 

\end{IEEEproof}
\textbf{Remarks}:
\begin{itemize}
    \item The proof does not conduct the mathematical induction on the tensor mode. This is because the uniqueness does not hold for matrix decomposition, i.e., $\min_{{\boldsymbol W}_1, {\boldsymbol W}_2}\|{\boldsymbol X} - {\boldsymbol W}_1{\boldsymbol G}{\boldsymbol W}_2^T\|_F$.
    \item This theorem is only a sufficient condition for the uniqueness of (\ref{tucker_opt}).
\end{itemize}

\begin{theorem}
Under the same condition as Lemma 3, using ALS to solve (\ref{tucker_opt}) by initializing the factor matrices with full rank, the achieved solution is unique when
\begin{align*}
\sum_{i\neq n} M_i \geq H_n.
\end{align*}
\end{theorem}
\begin{IEEEproof}
If we solve the optimization problem (\ref{tucker_opt}) through ALS as well as initializing factor matrices as full rank, updated factor matrices are with full rank at each iteration if we assume
\begin{align*}
\sum_{i\neq n} M_i  \geq H_n
\end{align*}
due to Lemma 2. Here, the prerequisite of Lemma 2 is met because in the updating rule for ${\boldsymbol W}_n$, $rank({\boldsymbol G}_{(n)}(\mathbf{W}_1 \otimes \cdots \mathbf{W}_{n-1}\otimes \mathbf{W}_{n+1}\otimes\cdots \otimes  \mathbf{W}_{N})^T) = H_n$ is guaranteed by the full rank initialization assumption on factor matrices. Therefore, when ALS terminated at an optimum of (\ref{tucker_opt}), we can assert this optimum is the unique by using Lemma 3.
\end{IEEEproof}
With minor revisions on the statement of the above Theorem, we can arrive at Theorem 1. 



\appendix[Low Complexity Factor Matrix Calculation in ALS]
At each step of using alternating least squares to solve the factor matrices ${\boldsymbol W}_{out-1}$ and ${\boldsymbol W}_{out-1}$ in (\ref{loss_function}), suppose we directly solve the following sub-problem to obtain ${\boldsymbol W}_{out-1}$,
\begin{align*}
    {\boldsymbol W}_{out-1} = \arg\min_{{\boldsymbol W}_{out-1}}\|{\boldsymbol Z}_{(1)} - {\boldsymbol W}_{out-1}{\boldsymbol G}_{(1)}( {\boldsymbol W}_{out-2}\otimes{\boldsymbol I}_{N_T}\otimes{\boldsymbol I}_{N_K})^T\|_F.
\end{align*}
The resulting memory costs spent on ${\boldsymbol W}_{out-2}\otimes{\boldsymbol I}_{N_T}\otimes{\boldsymbol I}_{N_K}$ are very large. To solve the bottleneck from memory units, we can alternatively solve the ${\boldsymbol W}_{out-1}$ as follows,
\begin{equation*}
\begin{aligned}
 {\boldsymbol W}_{out-1} &= \arg\min_{{\boldsymbol W}_{out-1}}\|{\boldsymbol {\mathcal Z}} - {{\boldsymbol {\mathcal G} }\times_1 {\boldsymbol W}_{out-1}\times_2 {\boldsymbol W}_{out-2}}\|\\
 &{\stackrel{(a)}{=}} \arg\min_{{\boldsymbol W}_{out-1}}\|{\boldsymbol {\mathcal Z}}-\sum_k {\boldsymbol {\mathcal G}^{(k)}}\times_1 {\boldsymbol W}_{out-1}^{(k)}\times_2{\boldsymbol W}_{out-2}^{(k)}\|  \\
&= \arg\min_{{\boldsymbol W}_{out-1}}\|{\boldsymbol { Z}}_{(1)}-\sum_k  {\boldsymbol W}_{out-1}^{(k)} ({\boldsymbol {\mathcal G}^{(k)}}\times_2{\boldsymbol W}_{out-2}^{(k)})_{(1)}\|\\
&= \arg\min_{{\boldsymbol W}_{out-1}}\|{\boldsymbol { Z}}_{(1)}- {\boldsymbol W}_{out-1}[({\boldsymbol {\mathcal G}^{(1)}}\times_2{\boldsymbol W}_{out-2}^{(1)})_{(1)}, \cdots, ({\boldsymbol {\mathcal G}^{(K)}}\times_2{\boldsymbol W}_{out-2}^{(K)})_{(1)}]^T \|
\end{aligned}
\end{equation*}
where $(a)$ comes from the partition definition of Tucker decomposition in (\ref{partition_tucker}) and $K := N!+1$. As the above calculation suggests, we can first calculate mode-2 product between each partitioned core tensor and factor matrix, then concatenate them as a big matrix to calculate a pseudoinverse to reach a least squares solution of ${\boldsymbol W}_{out-1}$. Similar tricks can be apply to calculate ${\boldsymbol W}_{out-2}$, $\cdots$, ${\boldsymbol W}_{out-N}$ in general multi-mode RC.


\ifCLASSOPTIONcaptionsoff
  \newpage
\fi



\bibliographystyle{IEEEtran}
\bibliography{IEEEabrv,ref.bib}
%

%







\end{document}